\documentclass{article}
\usepackage{color}
\usepackage[T2A]{fontenc}
\usepackage[english]{babel}
\usepackage{amssymb,amsmath}

\usepackage{graphicx}
\usepackage{epstopdf}
\usepackage{alltt}
\usepackage{multirow}
\usepackage[all]{xy}

\newcommand{\boxfigure}[1]%
        {\framebox[\textwidth]{%
         \parbox{0.95\textwidth}{#1}}}

\newtheorem%
     {theorem}{Theorem}[section]
\newtheorem%
     {corollary}[theorem]{Corollary}
\newtheorem%
     {proposition}[theorem]{Proposition} 
\newtheorem%
     {lemma}[theorem]{Lemma} 

\newtheorem%
     {exerciseAux}[theorem]{Exercise}
\newenvironment%
     {exercise}{\begin{exerciseAux}\rm}{\end{exerciseAux}}

\newtheorem%
     {beispielAux}[theorem]{Beispiel}
\newenvironment%
     {beispiel}{\begin{beispielAux}\rm}{\end{beispielAux}}

\newtheorem%
     {definitionAux}[theorem]{Definition}
\newenvironment%
     {definition}{\begin{definitionAux}\rm}{\end{definitionAux}}
\newtheorem%
     {remarkAux}[theorem]{Remark}
\newenvironment%
     {remark}{\begin{remarkAux}\rm}{\end{remarkAux}}

\newtheorem%
     {exampleAux}[theorem]{Example} 
\newenvironment%
   {example}{\begin{exampleAux}\parskip=0in\rm}{\parskip=0.1in\end{exampleAux}}

\newtheorem%
     {examplesAux}[theorem]{Examples} 
\newenvironment%
     {examples}{\begin{examplesAux}\rm}{\end{examplesAux}}

\newtheorem%
     {constructionAux}[theorem]{Construction} 
\newenvironment%
     {construction}{\begin{constructionAux}\rm}{\end{constructionAux}}

\def\qed{\hfill{\boxit{}}
  \ifdim\lastskip<\medskipamount \removelastskip\penalty55\medskip\fi}
\def\proofend{\hfill{\boxit{}}
  \ifdim\lastskip<\medskipamount \removelastskip\penalty55\medskip\fi}
\long\def\boxit#1{\vbox{\hrule\hbox{\vrule\kern3pt
                  \vbox{\kern3pt#1\kern3pt}\kern3pt\vrule}\hrule}}

\newenvironment%
     {genass}%
        {\medbreak\noindent{\bf General Assumption.\enspace}\it}%
        {\ifdim\lastskip<\medskipamount \removelastskip\penalty55\medskip\fi}

\newenvironment
     {comeqns}%
        {\vspace{-.5ex}      
         \[   %
              \begin{array}{lcl@{\qquad}rl}}%
        {\end{array}
         \]
         \vspace{-.5ex}}

\newenvironment
     {dbenum}%
        {\vspace{-.5ex}      
         \[   %
              \begin{array}{r@{\quad}l@{\qquad\qquad}r@{\quad}l}}%
        {\end{array}
         \]
         \vspace{-.5ex}}

\newcounter{zaehler}

\newenvironment
     {paireqns}%
        {\vspace{-.5ex}      
         \[   %
              \begin{array}{rcl@{\qquad}rcl}}%
        {\end{array}
         \]
         \vspace{-.5ex}}

\newcommand{\Op}{\mbox{\sl Op}}
\newcommand{\op}{\mbox{\sl op}}
\newcommand{\IN}{\mathbb{N}}
\newcommand{\Sol}{\mbox{\sl Sol}}
\newcommand{\Inf}{\mbox{\sl Subs}}
\newcommand{\Pref}{\mbox{\sl Pref}}
\newcommand{\Suff}{\mbox{\sl Suf}}
\newcommand{\Id}{\mbox{\sl Id}}
\newcommand{\keyword}[1]{\mbox{\bf{#1}}}
\newcommand{\myleft}[1]{\overleftarrow{#1}}
\newcommand{\myright}[1]{\overrightarrow{#1}}
\newcommand{\myleftright}[1]{\overleftrightarrow{#1}}
\newcommand{\leftEquiv}{\myleft{=}}
\newcommand{\rightEquiv}{\myright{=}}
\newcommand{\leftrightEquiv}{\myleftright{=}}
\newcommand{\leftA}{\myleft{\mathcal{A}}}
\newcommand{\rightA}{\myright{\mathcal{A}}}
\newcommand{\leftC}{\myleft{\mathcal{C}}}
\newcommand{\rightC}{\myright{\mathcal{C}}}
\newcommand{\leftrightC}{\myleftright{\mathcal{C}}}

\begin{document}

\title{Good parts first - a new algorithm for approximate search in lexica and string databases}
\author{Stefan Gerdjikov, Stoyan Mihov, Petar Mitankin, Klaus U. Schulz}
\date{}
\maketitle

\begin{abstract}
We present a new efficient method for approximate search in electronic lexica. Given an input string (the pattern) and a similarity threshold, the algorithm retrieves all entries of the lexicon
that are sufficiently similar to the pattern. Search is organized in subsearches that always start with an exact partial match where
a substring of the input pattern is aligned with a substring of a lexicon word.
Afterwards this partial match is extended stepwise to larger substrings. For aligning further parts of the pattern with corresponding parts of lexicon entries, more errors are tolerated at each subsequent step.
For supporting this alignment order, which may start at any part of the pattern,
the lexicon is represented as a structure that enables immediate access to any substring of a lexicon word and permits the extension of such substrings in both directions. Experimental evaluations of the approximate search procedure are given that show significant efficiency improvements compared to existing techniques. Since the technique can be used for large error bounds it offers interesting possibilities for approximate search in special collections of ``long'' strings, such as phrases, sentences, or book titles. 
\end{abstract}

\section{Introduction}

The problem of approximate search in large lexica is central for many applications like
spell checking, text and OCR correction \cite{Kukich92a,Dengel+97},
internet search \cite{Cucerzan+04,ahmad05,lau99},
computational biology \cite{Gusfield97} etc.
In a common setup the problem may be formulated as follows: A large set of words/strings called the lexicon is given as a static background resource. Given an input string (the pattern), the task is to efficiently find all entries of the lexicon where the Levenshtein distance between pattern and entry does not exceed a fixed bound specified by the user. The Levenshtein distance \cite{Levenshtein66} is often replaced by related distances. In the literature, the problem has found considerable attention, e.g. \cite{Kemal,YN98,BCP02,MS04}.


Classical solutions to the problem \cite{Kemal} try to align the pattern $P$ with suitable lexicon words in a strict left-to-right manner, starting at the left border of the pattern. The lexicon is represented as a trie or deterministic finite-state automaton, which means that each prefix of a lexicon word is only represented once and corresponds to a unique path beginning at the start state.
During the search, only prefixes of lexicon words are visited where the distance to a prefix $P'$ of the pattern does not exceed the given bound $b$. As a filter mechanism that checks if these conditions are always met, Ukonnen's method \cite{Uk85} or Levenshtein automata \cite{SM02} have been used.
The main problem with this solution is the so-called ``wall effect'': if we tolerate $b$ errors and start searching in the lexicon from left to right, then in the first $b$ steps we have to consider {\em all} prefixes of lexicon words. Eventually, only a tiny fraction of these prefixes will lead to a useful lexicon word, which means that our exhaustive initial search represents a waste of time.


In order to avoid the wall effect, we need to find a way of searching in the lexicon such that during the initial alignment steps between pattern and lexicon words the number of possible errors is  as small as possible. The ability to realize such a search is directly related to the way the lexicon is represented.  In \cite{MS04} we used two deterministic finite-state automata as a joint index structure for the lexicon. The first ``forward'' automaton represents all lexicon entries as before. The second ``backward'' automaton represents all reversed entries of the lexicon.  Given an erroneous input pattern, we distinguished two subcases: (i) most of the discrepancies between the pattern and the lexicon word are in the first half of the strings; and (ii) most of the discrepancies are in the second half. We apply two subsearches. For subsearch (i) we use the forward automaton. During traversal of the first half of the pattern we tolerate at most $b/2$ errors. Then search proceeds by tolerating up to $b$ errors. For subsearch (ii) the traversal is performed on the reversed automaton and the reversed pattern in a similar way -- in the first half starting from the back only $b/2$ errors are allowed, afterwards the traversal to the beginning tolerates $b$ errors. In \cite{MS04} it was shown that the performance gain compared to the classical solution is enormous and at the same time no candidate is missed.

In this paper we present a method that can be considered as an extension of the latter. The new method uses ideas introduced in the context of approximate search in strings in \cite{WuManber92,M94,BaezaYatesNavarro99,NavarroBaezaYates99,BaezaYatesNavaro00}. Assume that the pattern can be aligned with a lexicon word with not more than $b$ errors.
Clearly, if we divide the pattern into $b+1$ pieces, then at least one piece will {\em exactly} match the corresponding substring of a lexicon word in the answer set. In the new approach we first find the lexicon substrings that exactly match such a given piece of the pattern (``good parts first''). Afterwards we continue by extending this alignment, stepwise attaching new pieces on the left or right side.
For the alignment of new pieces, more errors are tolerated at each step, which guarantees that eventually $b$ errors can occur. Since at later steps the set of interesting substrings to be extended is already small the wall effect is avoided, it does not hurt that we need to tolerate more errors.
For this kind of search strategy, a new representation of the lexicon is needed where we can start traversal at any point of a word. 
In our new approach, the lexicon is represented as symmetric compact directed acyclic word graph (SCDAWG) \cite{BlumerBlumer87,Inenaga01} - a bidirectional index
structure where we
(i) have direct access to every substring of a lexicon word and
(ii) can deterministically extend any such substring {\em both to the left
and to the right} to larger substrings of lexicon words. This index structure can be
seen as a part of a longer development of related index structures \cite{BlumerBlumer87,Stoye95,Gusfield97,Breslauer98,Stoye04,Maass00,Inenaga01,Inenaga05,MMW09} extending 
work on suffix tries, suffix trees, and directed acyclic word graphs (DAWGs) \cite{Weiner73,McCreight76,Ukkonen95,ChenSeiferas84,BlumerBlumer85}.

Our experimental results show that the new method is much faster than previous methods mentioned above. For small distance bounds it often comes close to the theoretical limit, which is defined as a (in practice merely hypothetical) method where precomputed solutions are used as output and no search is needed. In our evaluation we not only consider ``usual'' lexica with single-word entries. The method is especially promising for collections of strings where the typical length is larger than in the case of conventional single-word lexica.
Note that given a pattern $P$ and an error bound $b$, long strings in the lexicon have long parts that can be exactly aligned with parts of $P$. This explains why even for large error bounds efficient approximate search is possible.  In our tests we used a large collection of book titles, and a list of 351,008 full sentences from MEDLINE abstracts as ``dictionaries''. In both cases, the speed up compared to previous methods is drastic. Future interesting application scenarios might include, e.g., approximate search in translation memories, address data, and related language databases. 

The paper is structured as follows. We start with some formal preliminaries in Section~\ref{sec:preliminaries}.
In Section~\ref{sec:example} we present our method informally using an example.
In Section~\ref{sec:searching} we give a formal description of the algorithm, assuming that an appropriate index structure for the lexicon with the above functionality is available.
In Section~\ref{sec:suffix_aut} we describe the symmetric compact directed acyclic word graph (SCDAWG).
Section~\ref{sec:evaluation} gives a detailed evaluation of the new method, comparing search times achieved with other methods. Experiments are based on various types of lexica, we also look at distinct variants of the Levenshtein distance.
In the Conclusion we comment on possible applications of the new method in spelling correction and other fields. We also add remarks on the historical sources for the index structure used in this paper.

\section{Technical Preliminaries}\label{sec:preliminaries}

{\em Words} over a given finite alphabet $\Sigma$ are denoted $P,U,V,W,\ldots$, symbols $\sigma,\sigma_i$ denote letters of $\Sigma$. The {\em empty word} is written $\varepsilon$. If $W= \sigma_1\cdots \sigma_n$, then $W^{rev}$ denotes the {\em reversed
word} $\sigma_n\cdots \sigma_1$. The $i$-th symbol $\sigma_i$ of the word $W = \sigma_1\cdots \sigma_n$ is denoted $W_i$.
In what follows the terms {\em string} and {\em word} are used interchangeably.
The {\em length} (number of symbols) of a word $W$ is denoted $\vert W\vert$.
We write $U\circ V$ or $UV$ for the {\em concatenation} of the  words $U,V\in \Sigma^\ast$.
A string $U$ is called a {\em prefix} (resp. {\em suffix}) of $W\in \Sigma^\ast$ iff $W$ can be represented in the form $W= U\circ V$ (resp. $W= V\circ U$) for some $V\in \Sigma^\ast$.
A string $V$ is a {\em substring} of $W\in \Sigma^\ast$ iff $W$ can be represented in the form $W= U_1\circ V \circ U_2$ for some $U_1, U_2\in \Sigma^\ast$.
The set of all strings over $\Sigma$ is denoted $\Sigma^{\ast}$, and the set of the nonempty strings over $\Sigma$ is denoted $\Sigma^{+}$.
By a {\em lexicon} or {\em dictionary} we mean a finite nonempty collection ${\mathcal{D}}$ of words. The set of all substrings (resp. prefixes, suffixes) of words in ${\mathcal{D}}$ is denoted $\Inf({\mathcal{D}})$ (resp. $\Pref({\mathcal{D}})$, $\Suff({\mathcal{D}})$).
The set of the reversed words from $\mathcal{D}$ is denoted $\mathcal{D}^{rev}$.
The {\em size of the lexicon} ${\mathcal{D}}$ is ${\vert}{\vert} {\mathcal{D}}\vert\vert := \sum_{W\in \mathcal{D}} \vert W\vert$.

\begin{definition}\label{DefinitionDFAS}
A {\em deterministic finite-state automaton} is a quintuple
$$
\mathcal{A}= (Q,\Sigma,s,\delta,F)
$$
where
$\Sigma$ is a finite
input alphabet, $Q$ is a finite set of states,
$s\in Q$ is the start state, $\delta: Q\times \Sigma \rightarrow Q$ is a partial transition function, and $F\subseteq Q$ is the set of final states.
\end{definition}

If $\mathcal{A}= (Q,\Sigma,s,\delta,F)$ is a deterministic finite-state automaton, the extended partial transition function $\delta^\ast$ is defined as usual: for each $q\in Q$ we have $\delta^\ast(q, \varepsilon) = q$. For a string $W\sigma$ ($W \in \Sigma^\ast, \sigma \in \Sigma$) $\delta^\ast(q, W\sigma)$ is defined iff both $\delta^\ast(q, W)=p$ and $\delta(p,\sigma)=r$ are defined. In this case, $\delta^\ast(q, W\sigma)=r$.
We consider  the size of a deterministic finite-state automaton $\mathcal{A}$ to be linear
in the number of states $\vert Q\vert$ plus the number of the transitions $\vert \{ (p,\sigma,q) \mid \delta(p, \sigma) = q \}\vert$.
Assuming that the size (number of symbols) of the alphabet $\Sigma$ is treated as a constant,
the {\em size} of $\mathcal{A}$ is $O(\vert Q\vert)$.

\begin{definition}\label{DefinitionGDFAS}
A {\em generalized deterministic finite-state automaton} is a quintuple
$\mathcal{A}= (Q,\Sigma,s,\delta,F)$, where $Q$, $\Sigma$, $s$ and $F$ are as above and
$\delta: Q\times \Sigma^+ \rightarrow Q$ is a partial function with the following property:
for each $q\in Q$ and each $\sigma \in \Sigma$ there exists at most one $U \in \Sigma^\ast$ such that $\delta(q, \sigma U)$ is defined.

\end{definition}
A transition $\delta(q,\sigma U) = p$ is called a {\em $\sigma$-transition} from $q$. The above condition then says that for each $q\in Q$ and each $\sigma \in \Sigma$ there exists at most one $\sigma$-transition. In what follows, $\sigma$-transitions of the above form are often denoted
$q \stackrel{\sigma U}{\rightarrow} p$. Let $${\cal V}= \{(q,V,p)\mid q,p\in Q, \mathcal{A} \mbox{ has a transition } q \stackrel{V}{\rightarrow} p\}.$$ The size of the generalized deterministic finite-state automaton $\mathcal{A}$ is considered to be $O(\vert Q\vert + \sum_{(q,V,p)\in {\cal V}} \vert V\vert)$, which is not $O(\vert Q\vert)$ in general.

\subsection{Suffix tries for lexica}\label{SubSecSuffixTries}

The following definitions capture possible index structures for search in lexica.
First, we define {\em the trie for a lexicon ${\mathcal{D}}$} as a tree-shaped deterministic finite-state automaton.
Each state of this automaton represents a unique prefix of lexicon words.
The final states represent complete words. Second, the {\em suffix trie for} ${\mathcal{D}}$ is defined as the trie of all suffixes in ${\mathcal{D}}$.

\begin{definition}\label{DefinitionTrie}
Let ${\mathcal{D}}$ be a lexicon over the alphabet $\Sigma$. The {\em trie for} ${\mathcal{D}}$ is the deterministic finite-state automaton
$Trie(\mathcal{D}) = (Q,\Sigma,q_{\varepsilon},\delta,\{q_{U} \mid U \in \mathcal D\})$ where $Q=\{q_U \mid U \in \Pref({\mathcal{D}})\}$ is a set of states indexed with the prefixes in $\Pref({\mathcal{D}})$ and $\delta(q_U,\sigma)=q_{U\circ \sigma}$ for all $U\circ \sigma \in \Pref({\mathcal{D}})$.
\end{definition}
Obviously, the size of $Trie(\mathcal{D})$ is $O(\vert\vert \mathcal{D}\vert\vert)$. While tries support left-to-right search for words of the lexicon, the next index structure supports left-to-right search for substrings of lexicon words.

\begin{definition}\label{DefinitionSuffixTrie}
Let ${\mathcal{D}}$ as above. The {\em suffix trie for} ${\mathcal{D}}$ is the deterministic finite-state automaton $STrie({\mathcal{D}}) := Trie(\Suff(\mathcal{D}))$.
\end{definition}
In general, the size of the suffix trie for ${\mathcal{D}}$ is $O(\vert \vert \Suff({\mathcal{D}}) \vert \vert)$ and
$\vert\vert \Suff({\mathcal{D}}) \vert\vert$ is quadratic with respect to $\vert\vert {\mathcal{D}} \vert\vert$.
For example, for every $n \in \mathbb{N}$ the number of states in $STrie(\{a^n b^n\})$ is $(n+1)^2$.\\

\noindent {\em Bidirectional suffix tries.\ }
We now introduce a bidirectional index structure supporting both left-to-right search and right-to-left search for substrings of lexicon words.
For $U\in \Sigma^\ast$ always $q_U$ is a state in $STrie({\mathcal{D}})$ iff $q_{U^{rev}}$ is a state in $STrie({\mathcal{D}}^{rev})$.
Hence, following Giegerich and Kurtz \cite{GiegerichKurtz97}, from the two suffix tries $STrie({\mathcal{D}})$ and $STrie({\mathcal{D}}^{rev})$ we obtain
one bidirectional index structure by identifying each pair of states $(q_U,q_{U^{rev}})$ from the two structures.

\begin{definition}\label{DefinitionBiSTrie} The {\em bidirectional suffix trie} for ${\mathcal{D}}$ is the tuple
$BiSTrie({\mathcal{D}}) := (Q, \Sigma, q_{\varepsilon}, \delta_L, \delta_R, F)$, where
$(Q, \Sigma, q_{\varepsilon}, \delta_R, G') = STrie({\mathcal{D}})$,
$F := \{ q_U \in Q \mid U \in {\mathcal{D}}\}$ and
$\delta_L : Q \times \Sigma \rightarrow Q$ is the partial function
such that
$(Q^{rev}, \Sigma, q_{\varepsilon}, \delta_L^{rev}, G'') = STrie({\mathcal{D}}^{rev})$ for
$Q^{rev} = \{ q_{U^{rev}} \mid q_U \in Q \}$ and
$\delta_L^{rev}(q_{U^{rev}}, x) = \delta_L(q_U, x)$.
\end{definition}

\begin{example}
The bidirectional suffix trie for ${\mathcal{D}} = \{ {\tt ear}, {\tt lead}, {\tt real} \}$, is shown in Figure~\ref{fig:bistrie}.
\end{example}
As in the case of one-directional structures, the main problem is the size of the index. In general, the size of $BiSTrie({\mathcal{D}})$ is quadratic in the size of the ${\mathcal{D}}$. The final structure, which will be presented in Section~\ref{sec:suffix_aut}, can be considered as a compacted version of the bidirectional suffix trie.

\begin{remark} It is known that a suffix tree for a lexicon $\mathcal{D}$ can be stored in space $O(\vert\vert \mathcal{D}\vert\vert)$
and built online in time $O(\vert\vert D\vert\vert)$, \cite{Ukkonen95}. Suffix trees are compacted variants of suffix tries.
In this paper we use compact directed acyclic word graphs \cite{BlumerBlumer87,Inenaga05} which are minimized variants of suffix trees.
\end{remark}

\subsection{Approximate search in lexica and Levenshtein filters}

\begin{definition}
The {\em Levenshtein distance} between $V,W\in \Sigma^\ast$, denoted $d_L(V,W)$, is the minimal number of edit operations needed to transform $V$ into $W$.
 {\em Edit operations} are the deletion of a symbol, the insertion of a symbol, and the substitution of a symbol by another symbol in $\Sigma$.
\end{definition}

\noindent
In what follows, $\Id := \{\langle \sigma,\sigma\rangle \mid \sigma\in \Sigma\}$ is considered as a set of identity operations.

\begin{definition}\label{DefGenDistance}
A {\em set of generalized weighted operations} is a pair $(\Op,w)$ where
\begin{enumerate}
\item
$\Op \subseteq \Sigma^\ast \times \Sigma^\ast$ is a finite set of operations such that $\Id\subseteq \Op$,
\item
$w: \Op\rightarrow {\IN}$ assigns to each operation $\op \in \Op$  a nonnegative integer weight
$w(\op)$ such that $w(\op) = 0$ iff $\op\in \Id$.
\end{enumerate}
If $\op = \langle X,Y\rangle$ represents an operation in $\Op$,
then $l(\op)$, {\em the left side of the operation}, is defined as $l(\op) = X$
and $r(\op)$, {\em the right side of the operation}, is defined as $r(\op) = Y$.
The {\em width} of $\op \in \Op$ is $\vert l(\op)\vert$.
\end{definition}


\begin{definition}\label{DefAlignment}
Let $(\Op,w)$ be a set of generalized weighted operations. An {\em alignment}  is an arbitrary
sequence $\alpha = \op_1 \op_2 \dots \op_n\in \Op^\ast$ of operations $\op_i \in \Op$. The notions of left (right) side and weight are extended to alignments in a natural way:
\begin{eqnarray*}
& l(\alpha)=l(\op_1)l(\op_2)\dots l(\op_n) \\
& r(\alpha)=r(\op_1)r(\op_2)\dots r(\op_n) \\
& w(\alpha)=\sum_{i=1}^n w(\op_i).
\end{eqnarray*}
\end{definition}

Note that Definition~\ref{DefAlignment} does not permit overlapping of operations in the sequence.
In our setting, operations that transform the left side into the right side are applied simultaneously.
Formally, each sequence of operations representing an alignment is a string over the alphabet $\Op$.

\begin{definition}\label{DefinitionDDistance}
The generalized distance induced by a given set of generalized weighted operations $(\Op,w)$
is the function $d:\Sigma^\ast \times \Sigma^\ast\rightarrow \IN \cup \{\infty \}$ which is defined as:
\begin{equation*}
d(V,W)=\min \{w(\alpha) \mid \alpha \in \Op^\ast, l(\alpha)=V \text{ and } r(\alpha)=W\}.
\end{equation*}
We say that $\alpha\in \Op^\ast$ is an {\em optimal alignment of $V$ and $W$} iff
$l(\alpha)=V$, $r(\alpha)=W$ and $w(\alpha)=d(V,W)$.
\end{definition}

\begin{remark}
In terms of Definition~\ref{DefinitionDDistance} we can represent the Levenshtein $d_L$ as
the distance induced by  $(\Op_L,w_L)$ where $\Op_L=(\Sigma\cup \varepsilon)\times (\Sigma\cup \varepsilon)\setminus\{\langle\varepsilon,\varepsilon \rangle\} $
and $w_L(\op)=1$ for all $\op\not \in \Id$.
\end{remark}

\begin{remark}
Given a set of generalized weighted operations $(\Op,w)$, dynamic programming can be used to efficiently compute $d(V,W)$
for strings $V$ and $W$, \cite{Uk85,Veronis88}.
\end{remark}


In this paper, we are interested in solutions for the following algorithmic problem ({\em ``approximate search in lexica''}):

{\em Let $\mathcal{D}$ be a fixed lexicon, let $d$ denote a given generalized distance between words. 
For an input pattern $P\in \Sigma^\ast$ and a bound $b\in \mathbb{N}$}, efficiently find all words $W\in \mathcal{D}$ such that $d(P,W)\leq b$.


\begin{definition}
Let $b\in \mathbb{N}$ denote a given bound. By a {\em Levenshtein filter} for bound $b$ we mean any  algorithm that takes as input two
words $P,U\in \Sigma^\ast$ and decides
\begin{enumerate}
\item\label{DPprob1}
if there exists a string
$V \in \Sigma^\ast$ such that $d_L(P,U\circ V) \leq b$,
\item\label{DPprob2}
if $d_L(P,U) \leq b$.
\end{enumerate}
More generally, if
$d$ is any generalized distance, a {\em filter for $d$} for bound $b$ is an
 algorithm that takes as input two
words $P,U\in \Sigma^\ast$ and decides
\begin{enumerate}
\item\label{FilterProblem}
if there exists a string
$V \in \Sigma^\ast$ such that $d(P,U\circ V) \leq b$,
\item\label{FilterProblem2}
if $d(P,U) \leq b$.
\end{enumerate}
Note that a filter for $d$ for bound $b$ does not depend on the lexicon $\mathcal{D}$.
\end{definition}

The interest in filters of the above form relies on the observation that
in approximate search in lexica we often face a given input pattern $P\in \Sigma^\ast$.
When we traverse the lexicon, which is represented as a trie or automaton, we
want to recognize at the earliest possible point if the current path, which represents a prefix $U$
of a lexicon word, can {\em not} be completed to any word that is close enough to $P$ (Decision Problem~\ref{DPprob1}). When reaching a final state representing a word $W=U$ of the lexicon we want to check if $W$ satisfies the bound (Decision Problem~\ref{DPprob2}).
In \cite{Kemal}, the matrix based dynamic programming approach was used to realize a Levenshtein filter. In \cite{SM02} we introduced the concept of a Levenshtein automaton, which represents a more efficient filter mechanism.

In what follows we make a more general use of filters. Our lexicon traversal below starts from a {\em substring} of a lexicon word, which is compared to a substring $P$ of the pattern. In addition to steps where we extend substrings on the right using a filter of the above form, we also use steps where we extend substrings with new symbols on the left. In this situation we need to check for given
$P,U\in \Sigma^\ast$ if there exists a string
$V$ such that $d(P,V\circ U) \leq b$. This means that with suitable extensions of $U$ on the left we might reach an interesting alignment partner for $P$ among the substrings of lexicon words.
%
%

\begin{remark}\label{RemarkReversedFilter}
Assume that we have an algorithm that, given a distance $d$  induced by $(\Op,w)$ and a bound $b$, constructs a filter for extension steps on the right of the above form. We may build a second filter for the symmetric distance $d^{rev} := (\{\op^{rev} \mid \op \in \Op\},w^{rev})$ where $\op^{rev} :=(l(\op)^{rev},r(\op)^{rev})$ and $w^{rev}(\op^{rev}) := w(\op)$ for $\op \in \Op$.
Obviously, for given $P,U\in \Sigma^\ast$ there exists a string
$V$ such that $d(P,V\circ U) \leq b$ iff there exists a string
$V'$ such that $d^{rev}(P^{rev},U^{rev}\circ V') \leq b$.
Hence the second ``reversed filter'' can be used to control
extension steps on the left.
\end{remark}

The use of filters is directly related to the ``wall effect''. When the lexicon offers many possibilities for extending a given prefix or substring of a lexicon word, then the search space in a crucial way depends on the bound $b$ of the filter that is used. When using  a large bound, a large number of extensions has to be considered. Note that typically short prefixes/substrings have a very large number of extensions in the lexicon, while long prefixes/substrings often point to a unique entry.
From this perspective, the problem discussed in the paper can be rephrased: we are interested in a search strategy where the use of large bounds in filters is only necessary for large substrings at the end of the search. When we construct alignments between the pattern and lexicon words, we want to build ``good parts'' first.


\section{Basic Idea}\label{sec:example}

In this section we explain the idea of our algorithm using a small example.
We also characterize the kind of resources needed to achieve its efficient implementation.
Consider the dictionary
\begin{equation*}
{\mathcal{D}}=\{{\tt ear}, {\tt real}, {\tt lead}\}.
\end{equation*}
Suppose that for the pattern
\begin{equation*}
P = {\tt dread}
\end{equation*}
we want to find all words $W$ in ${\mathcal{D}}$ such that $d_L(P,W)\leq 2$.
The standard way to solve the problem is a left-to-right search in the lexicon, using a filter for bound $2$.
As described above, we want to avoid the use of a large filter bound at the beginning of the search.
We next illustrate a first approach along  these lines, which is then refined.

Let $W$ in ${\mathcal{D}}$ such that $d_L(P,W)\leq 2$. When we split $P={\tt dread}$ into the
three parts ${\tt d}$, ${\tt re}$, ${\tt ad}$, then there must be a corresponding representation of
$W$ in the form $W=W_1\circ W_2\circ W_3$ such that $d_L({\tt d},W_1)+ d_L({\tt re},W_2) + d_L({\tt ad},W_3) \leq 2$.
We distinguish three cases, $d_L({\tt d},W_1) = 0$,  $d_L({\tt re},W_2) = 0$, or  $d_L({\tt ad},W_3) = 0$. This leads to the following three subtasks:
\begin{enumerate}
\item\label{Subtask1}
Check if ${\tt d}$ represents a substring of a word in ${\mathcal{D}}$. In the positive case, look for extensions $V$ of ${\tt d}$ on the right to words of the form ${\tt d}\circ V\in {\mathcal{D}}$ such that $d_L({\tt dread},{\tt d}V)\leq 2$.
\item\label{Subtask2}
Check if ${\tt re}$ represents a substring of a word in ${\mathcal{D}}$. In the positive case, look for extensions $V_2$ of ${\tt re}$ on the right and  extensions $V_1$ of ${\tt re}V_2$ on the left to words of the form
$V_1 \circ {\tt re} \circ V_2\in {\mathcal{D}}$ such that $d_L({\tt dread},V_1{\tt re}V_2)\leq 2$ .
\item\label{Subtask3}
Check if ${\tt ad}$ represents a substring of a word in ${\mathcal{D}}$. In the positive case, look for extensions $V$ of ${\tt ad}$ on the left to words of the form
$V \circ {\tt ad}\in {\mathcal{D}}$ such that $d_L({\tt dread},V{\tt ad})\leq 2$.
\end{enumerate}

\begin{figure}
\begin{center}
\includegraphics[width=0.75\textwidth]{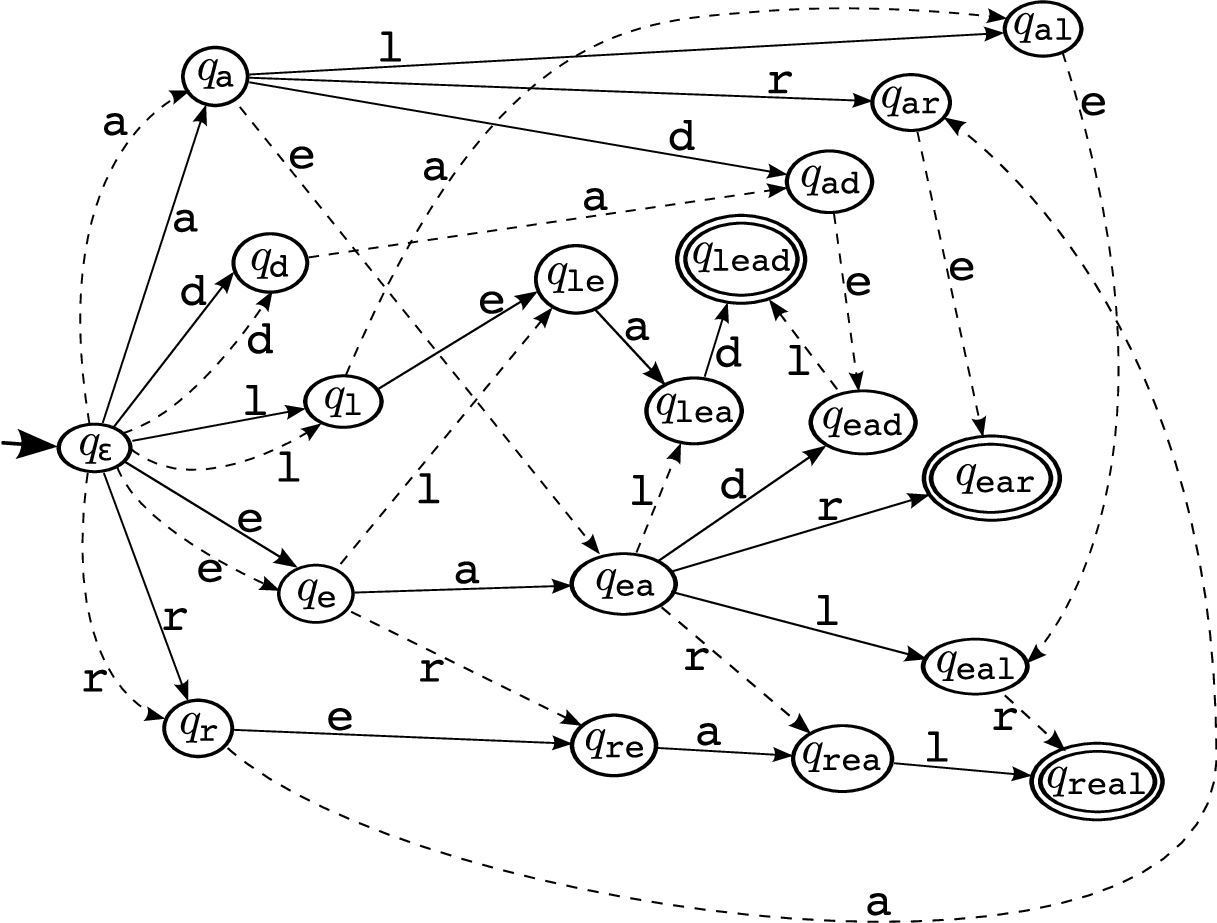}
\caption{The bidirectional suffix trie $BiSTrie({\mathcal{D}})$ for ${\mathcal{D}} = \{ {\tt ear}, {\tt lead}, {\tt real} \}$
represents all substrings of ${\mathcal{D}}$ and allows extending each substring either to the right
by following the \emph{solid arcs} or to the left by following the \emph{dashed arcs}.}
\label{fig:bistrie}
\end{center}
\end{figure}

The above task can be solved using an appropriate bidirectional index structure.
As an illustration\footnote{We should stress that $BiSTrie({\mathcal{D}})$ is just used for illustration purposes.
In general, the size of $BiSTrie({\mathcal{D}})$ is quadratic in the size of the ${\mathcal{D}}$,
which means that a more condensed structure is needed in practice.} we use the
bidirectional suffix trie (cf. Def.~\ref{DefinitionBiSTrie}) for ${\mathcal{D}} = \{ {\tt ear}, {\tt lead}, {\tt real} \}$, which is shown in Figure~\ref{fig:bistrie}.
The nodes $q_{U}$ of the graph depicted correspond to the substrings $U$ of our lexicon ${\mathcal{D}}$,
nodes marked with a double ellipse represent words in ${\mathcal{D}}$.
Following the \emph{solid arcs} we extend the current substring to
the right. Starting from $q_{\varepsilon}$ and traversing solid arcs we find any substring. If we follow the \emph{dashed arcs} we extend the current substring to the left.

It should be obvious how we may use the graph to solve the three subtasks in our example mentioned above.
As an example, we consider
Subtask~\ref{Subtask2}. Using the index we see that ${\tt re}$ is a substring of a word in ${\mathcal{D}}$.  Right extension steps of ${\tt re}$ in the index are controlled using a Levenshtein filter
for pattern suffix ${\tt read}$ and bound $2$. We find the two extensions ${\tt rea}$ and ${\tt real}$. Then, for the left extension steps  we use the filter for the full pattern ${\tt dread}$ and bound $2$. The index shows that both ${\tt rea}$ and ${\tt real}$ cannot be extended on the left. However, since already $d_L({\tt dread},{\tt rea}) \leq  2$ and $d_L({\tt dread},{\tt real}) \leq  2$ the filter licenses the empty left extension.
Among the two resulting substrings, ${\tt real} \in {\mathcal{D}}$ is a solution.
In a similar way, solving Subtask~\ref{Subtask3} leads to the second solution ${\tt lead}$.
When we abstract from our small example, the above procedure gives rise to the following

{\bf First search idea.\ }
Split $P=P_1\circ \cdots\circ P_{b+1}$ into $b+1$ parts $P_i$ of approximately the same length and apply $b+1$ subsearches.
For the $i$-th subsearch, first check if $P_i$ is a substring of a lexicon word (Step 1). In the positive case, try to extend $P_i$ to larger substrings of lexicon words, using a Levenshtein filter for bound $b$ (Step 2). 

A nice aspect of this a search strategy is that each subsearch starts with an exact match (Step 1), which represents a search with filter bound $0$. 
However, afterwards in Step 2 we immediately use a Levenshtein filter for the full bound $b$ for all left and right extension steps. If $b$ is large, this may lead to a large search space.

{\bf Improved search idea.\ } We now look for a refinement where we can use small filter bounds for the initial extension steps.
To this end, we first slightly generalize the problem and search all \emph{substrings} $V$ of words in ${\mathcal{D}}$ such that $d_L(P,V)\leq b$.
Afterwards we simply filter those substrings that represent entries of ${\mathcal{D}}$.

\begin{figure}
\begin{center}
\includegraphics[width=0.4\textwidth]{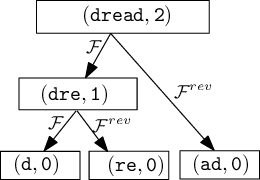}
\caption{Reducing the original query $({\tt dread},2)$ into simpler ones. As a result we obtain an ordered  binary tree representing search alternatives. The labels of the arcs indicate what sort of filter has to be used at the extension steps. The label $\mathcal{F}$ shows that
we extend to the right and thus an ordinary filter is required, whereas the label $\mathcal{F}^{rev}$ means that we extend to the left and
therefore a reverse filter (see Remark~\ref{RemarkReversedFilter}) has to supervise this step. The bound that determines a filter
coincides with the threshold of the query written in the parent node.}
\label{fig:query_tree_filters}
\end{center}
\end{figure}

We illustrate the improved search procedure using again our small example.
In what follows, the notation $({\tt dread},2)$ is used as a shorthand for the algorithmic task to find all substrings $V \in \Inf({\mathcal{D}})$ such that $d_L({\tt dread},V)\leq 2$, and similarly for other strings and bounds. The expression $({\tt dread},2)$ is called a {\em query} with query pattern ${\tt dread}$ and bound $2$.
Now consider the query tree depicted in Figure~\ref{fig:query_tree_filters}.
%
%
%
%
The idea is to solve the problems labeling the nodes in a bottom-up manner.
The three leaves exactly correspond to
the Steps 1 in the three subtasks discussed above: in fact, to solve the problems $({\tt d},0)$, $({\tt re},0)$ and $({\tt ad},0)$ just means to check if
${\tt d}$, ${\tt re}$, or ${\tt ad}$ are substrings of lexicon words.
We then solve problem $({\tt dre},1)$. This involves two independent steps.
\begin{enumerate}
\item
We look for extensions of the substring ${\tt d}$ (as a solution of the left child in the tree) at the right.
\item
We look for extensions of the substring ${\tt re}$ (as a solution of the right child in the tree) at the left.
\end{enumerate}

It is important to note that both  extension steps are controlled using a Levenshtein filter for bound $1$ for $P' = {\tt dre}$ (see Figure~\ref{fig:query_tree_filters}). As a result we obtain the single solution ${\tt re}$ for the query $({\tt dre},1)$. The next step in the bottom-up procedure looks at the root node $({\tt dread},2)$. Solving this node again involves two independent steps.
\begin{enumerate}
\item
We look for extensions of the substring ${\tt re}$ (as a solution of the left child in the tree) at the right.
\item
We look for extensions of the substring ${\tt ad}$ (as a solution of the right child in the tree) at the left.
\end{enumerate}
At this final step we cannot avoid the use of a Levenshtein filter for ${\tt dread}$ and bound $2$. We respectively obtain (1) ${\tt rea}$, ${\tt real}$ and (2) ${\tt dead}$, ${\tt lead}$.


Comparing the two search strategies, we see that at least Subtasks 1 and 2 have been replaced by a subsearch where we use filter bound $2$ only at the last extension step where we already found ${\tt re}$ and want to solve $({\tt dread},2)$ adding right extensions. More generally, search trees of this form offer a possibility to postpone the use of large filter bounds to the end of the search. Details will be given in the next section where we formally describe the refined procedure.

\begin{remark}\label{RemarkIndexFilterNeeded}
In order to efficiently realize a bottom-up subsearch of the form indicated above we need
\begin{enumerate}
\item
an index structure that supports the following tasks:
\begin{enumerate}
\item\label{task1}
given a string $V$, efficiently decide if $V$ represents a substring of a lexicon word,
\item\label{task2}
given a substring $V$ of a lexicon word, give immediate access to all substrings of lexicon words of the form $V\circ \sigma$ that add one letter $\sigma\in \Sigma$ to the right,
\item\label{task3}
given a substring $V$ of a lexicon word, give immediate access to all substrings of lexicon words of the form $\sigma \circ V$ that add one letter $\sigma\in \Sigma$ to the left.
\end{enumerate}
\item
A filter for the bound $b$ specified at the parent node faced at an upward step. The filter takes as first input the query pattern $P'$ specified at the parent node. Subsearches start with a given solution of the left (right) child query.
When adding letters to the right (left) we use a conventional (``reversed'') filter, cf. Remark~\ref{RemarkReversedFilter}.
\end{enumerate}
\end{remark}

\begin{remark}
A similar idea was introduced by Navaro and Baeza-Yates, \cite{BaezaYatesNavaro00}, for approximate search of a pattern in the set of substrings of a long text. In \cite{BaezaYatesNavaro00} the authors use suffix arrays for its realization and analyze
how to organize the splitting to optimize the efficiency of this approach in terms of the length of the text and pattern. Their theoretical results show that this technique improves over the naive algorithm in some cases, but still it does not avoid the wall effect in general.
In \cite{NavarroBaezaYates99} the same authors present an algorithm for online approximate search of substrings of a long text.
Their algorithm, as the algorithm presented here, uses binary trees representing search alternatives to reduce the search space.
The essential difference is that their algorithm is online, i.e. does not rely on a precomputed index.
\end{remark}

\section{Search procedure}\label{sec:searching}

The purpose of this section is to provide a formal description of the approach considered
in the previous section.
In what follows we assume that ${\mathcal{D}}$ is a fixed lexicon and $d$ is a given generalized distance induced by $(\Op,w)$.
For input strings $P\in \Sigma^\ast$ and a bound $b\in \mathbb{N}$  we want
to retrieve all words $W\in {\mathcal{D}}$ such that $d(P,W)\le b$.
We consider the case where each operation $op\in Op$ has width $\leq 1$ (cf. Def.~\ref{DefGenDistance}).
In the Appendix we show how essentially the same technique can be used for arbitrary generalized
distances.

\begin{definition}
A {\em query} is a pair $(P',b')$ where $P'\in \Sigma^\ast$ is a substring of $P$ and $b' \leq b$.
The set $\Sol_{D}(P',b') := \{V\in \Inf({\mathcal{D}}) \mid d(P',V)\le b'\}$ is called the {\em solution set} for $(P',b')$.
\end{definition}

The search procedure has three phases. We first build a search tree for
the query $(P,b)$. Then, using a bottom-up procedure we solve
all queries of the search tree, in particular $(P,b)$. The final step is trivial. We simply select from $\Sol_{D}(P,b)$ those elements that represent entries of ${\mathcal{D}}$.

\subsection{Building the search tree for a pattern}


We explain how to obtain for a given pattern $P$ a binary tree ${\mathcal T}_P$ with queries assigned to each node, see Figure~\ref{fig:query_tree_filters}.

Select any rooted ordered tree ${\mathcal T}$ with $b+1$ leaves $\lambda_1,\ldots,\lambda_{b+1}$ (enumerated in canonical left-to-right ordering)
where each non-leaf node has exactly $2$ children.
Then decorate the nodes of ${\mathcal T}$ with queries to define the search tree ${\mathcal T}_P$:
Split the pattern $P$ in the form $P=P_1\circ P_2\dots \circ P_{b+1}$ where the $P_i$ are substrings of $P$
of almost equal length, i.e. $||P_i|-|P_j||\le 1$ ($1 \leq i,j \leq b+1$). To each leaf $\lambda_i$ of ${\mathcal T}$ assign the query
$(P_i,0)$.  To each non-leaf node $\eta$ of ${\mathcal T}$ assign the query $(P_i\circ \cdots \circ P_{i+b'},b')$ where $\lambda_i,\ldots,\lambda_{i+b'}$
is the sequence of leaves representing descendants of $\eta$ in ${\mathcal T}$ in the natural left-to-right ordering.
(Note that the root of ${\mathcal T}_P$ has label $(P,b)$, which is the original query.)

\begin{example}
In the example considered in Section~\ref{sec:example}  we had $b=2$, $P={\tt dread}$, $P_1={\tt d}$, $P_2={\tt re}$, $P_3={\tt ad}$.
As our starting point ${\mathcal T}$ for decoration, we selected one among two possible binary rooted ordered trees.
\end{example}

\begin{remark}
The choice of a tree ${\mathcal T}$ satisfying the above conditions influences the time needed to solve the query. The general philosophy is to
avoid queries of the form $(P',b')$ where $P'$ is a short word and $b'$ is a large bound. A good choice is the use of a balanced tree where all paths from the root reach a certain length. Other optimizations represent a possible subject for future studies.
\end{remark}

\subsection{Computation of solution sets}\label{subsectionComputationSolutionsSets}


For each query $(P',b')$ of the tree ${\mathcal T}_P$ we compute a set $S_D(P',b')$ in a bottom-up fashion. We shall prove below that $S_D(P',b')$ is the solution set $\Sol_D(P',b')$ in each case.

{\em Initialization steps.\ } For a leaf query $(P_i,0)$  we decide if $P_i$ is a substring of a lexicon word. In the positive case we let $S_D(P_i,0) := \{P_i\}$, otherwise we define $S_D(P_i,0) := \emptyset$.

\begin{definition}\label{def:extension-steps} {\em Extension steps.\ }
Let $(P',b')$ denote the query at a non-leaf node $\eta$ of ${\mathcal T}_P$,
let $(P_1',b_1')$ and $(P_2',b_2')$ denote the queries of the two children $\eta_1, \eta_2$ of $\eta$, which are given in the natural left-to-right ordering.
Given the sets $S_D(P_1',b_1')$ and $S_D(P_2',b_2')$ we define $S_D(P',b')$ as the union of the two sets
$S_1$ and $S_2$ defined as
\begin{eqnarray*}
S_1&:=&\{U \circ V\in \Inf({\mathcal{D}}) \mid U\in S_D(P_1',b_1'), d(P_1',U)+d(P_2',V)\le b'\}\\
S_2&:=&\{V \circ U\in \Inf({\mathcal{D}}) \mid U\in S_D(P_2',b_2'), d(P_1',V)+d(P_2',U)\le b'\}
\end{eqnarray*}
\end{definition}

\begin{proposition}\label{PropositionCorrectness}
The computation of solution sets is correct: for each query $(P',b')$ of ${\mathcal T}_P$ we have $S_D(P',b')=\Sol_D(P',b')$.
\end{proposition}

\subsection{Correctness proof and remarks}

To prove Proposition~\ref{PropositionCorrectness},  some preparations are needed.

\begin{remark}\label{RemarkOnTree}
Let $\eta$ denote a non-leaf node of ${\mathcal T}_P$ decorated with query $(P',b')$. Let $(P_1',b_1')$ and $(P_2',b_2')$ denote the queries of the two children $\eta_1, \eta_2$ of $\eta$, which are given in the natural left-to-right ordering. Then we have $P'= P_1'\circ P_2'$ and $b_1'+b_2'=b'-1$.
\end{remark}






\begin{proposition}\label{PropositionSplitAlternative}
Let ${\mathcal{D}}$ and $d= (\Op,w)$ as above, assume that each operation in $\Op$ has width $\leq 1$, let $b\in \IN$. If
$P\in \Sigma^\ast$ is a word with $P=P_1\circ P_2$ and $\alpha\in \Op^\ast$ is an alignment with $l(\alpha)=P$ and weight $w(\alpha)\leq b$, then
\begin{enumerate}
\item
$\alpha$ can be represented in the form $\alpha=\alpha_1\circ \alpha_2$ such that $l(\alpha_1)=P_1$ and $l(\alpha_2)=P_2$.
\item
if $b'$ and $b''$ are two arbitrary integers with the property $b'+b''=b-1$, then
$w(\alpha_1)\le b'$ or $w(\alpha_2)\le b''$.
\end{enumerate}
\end{proposition}

{\em Proof.\ }
 Since $\alpha$ is a sequence of operations $\alpha=op_1\dots op_n$ and $P=l(op_1)\dots l(op_n)$, the first
part follows immediately from the fact that $|l(op_i)|\le 1$ ($1\leq i\leq n$).
The second statement is an obvious consequence.
\qed

\begin{corollary}\label{CorollarySplit}
If $P$ and $W$ are arbitrary words with $d(P,W)\le b$ and $P=P_1\circ P_2$, then:
\begin{enumerate}
\item $W$ can be represented in the form $W=W_1\circ W_2$ such that $d(P,W)=d(P_1,W_1) + d(P_2,W_2)$.
\item if $b'+b''=b-1$, then $d(P_1,W_1)\le b'$ or $d(P_2,W_2)\le b''$,
\end{enumerate}
\end{corollary}

{\em Proof.\ } 
 Let $\alpha$ be an optimal alignment
of $P$ and $W$. Then $w(\alpha) = d(P,W)\leq b$. We can define $W_i=r(\alpha_i)$ for $i=1,2$ where $\alpha_i$ are the alignments provided by Proposition~\ref{PropositionSplitAlternative}. The second statement follows
since $w(\alpha_i)\ge d(P_i,W_i)$, by the definition of a $d$-distance.
\qed

{\bf (Proof of Proposition~\ref{PropositionCorrectness}.)\ } This is obvious for the leaf queries. Consider a non-leaf node
$\eta$ of ${\mathcal T}_P$ with query  $(P',b')$, let $(P_1',b_1')$ and $(P_2',b_2')$ denote the queries of the two children $\eta_1, \eta_2$ of $\eta$, which are given in the natural left-to-right ordering. We may assume that $S_D(P_i',b_i')=\Sol_D(P_i',b_i')$ for $i=1,2$. Remark~\ref{RemarkOnTree} shows that $P' = P_1'\circ P'_2$.
Consider an element $U \circ V$ of $S_1$ (we use the notation introduced in Section~\ref{subsectionComputationSolutionsSets}). We have
$d(P',U \circ V) = d(P_1'\circ P'_2,U \circ V) \leq d(P_1',U)+d(P_2',V) \leq b'$ which shows that
$U \circ V \in \Sol_D(P',b')$. Hence $S_1\subseteq \Sol_D(P',b')$.
Similarly we see that $S_2\subseteq \Sol_D(P',b')$.
Conversely consider an element $W \in \Sol_D(P',b')$. Since $P' = P_1'\circ P'_2$, Corollary~\ref{CorollarySplit} shows that $W$ can be represented as $W=W_1 \circ W_2$
such that $d(P'_1,W_1)+d(P'_2,W_2)=d(P',W)$ and we have
(1) $d(P_1',W_1)\le b_1'$ or (2) $d(P'_2,W_2)\le b_2'$. In case (i), $W_i \in S_D(P_i',b_i')=\Sol_D(P_i',b_i')$, which shows that $W= W_1\circ W_2$ is found in $S_i$ ($i=1,2$).
Hence $\Sol_D(P',b') \subseteq S_1 \cup S_2 = S_D(P',b')$.
\qed

\begin{remark}
It is simple to see that in the example presented in the previous section the computation of solutions sets
follows exactly the above procedure.
Following the definition of the Extension steps, we need to
construct the complete sets of candidates $S_D(P_1',b_1')$ and $S_D(P_2',b_2')$
in order to compute the sets $S_1$ and $S_2$ for the query node $(P',b')$
and eventually determine $S_D(P',b')=S_1 \cup S_2$. This corresponds to
a bottom-up  traversal of the search tree $\mathcal{T}_P$.
%
%
\end{remark}

\begin{remark}\label{remarkIndependentExtension}
Observe that given a candidate $U_1\in S_D(P_1',b_1')$, the set of
successful candidates $U_1\circ V_1\in S_1$ with $d(U_1,P_1')+d(V_1,P_2')\le b'$
which result in the right extension steps depend only on the specific
candidate $U_1$, the dictionary ${\cal D}$ and the query node $(P',b')$ but not
on $S_D(P_1',b_1')$.
A similar observation is valid for the successful candidates $U_2\in S_D(P_2',b_2')$.
\end{remark}

\begin{remark}\label{remarkDFSTraversalIsPossible}
Remark~\ref{remarkIndependentExtension} means that the target set $S(P,b)$ can
 be constructed by using any traversal algorithm $\mathcal{A}$ of the search tree $\mathcal{T}(P)$ which satisfies
 the following three conditions:
 \begin{enumerate}
 \item it correctly initializes the candidate sets $S_D(P_i,0)$ for each leaf $(P_i,0)$.
 \item if $\mathcal{A}$ generates a candidate $U_1$ in a node $(P_1',b_1')$ which is a left child
 of $(P',b')$, then $\mathcal{A}$ generates also all successful candidates $U_1\circ V_1$ for
 the node $(P',b')$ such that $d(U_1,P_1') + d(V_1,P_2')\le b'$.
 \item if $\mathcal{A}$ generates a candidate $U_2$ in a node $(P_2',b_2')$ which is a right child
 of $(P',b')$, then $\mathcal{A}$ generates also all successful candidates $U_1\circ V_1$ for
 the node $(P',b')$ such that $d(U_1,P_1') + d(V_1,P_2')\le b'$.
 \end{enumerate}
 In particular one can replace the bottom-up traversal by a depth first search algorithm.
\end{remark}

\begin{remark}\label{remarkIndexStructure}
It is obvious to see that the efficient realization of the above search algorithm
can be based on the resources described in Remark~\ref{RemarkIndexFilterNeeded}. The efficient computation of the sets $S_1$ and $S_2$ in the
bottom-up steps is achieved by using the given index structure for extensions on the right and left, respectively. Each extension by a single letter is controlled using a filter for the
generalized distance $d$ for the appropriate bound.
As we mentioned earlier, the index structure
shown in Figure~\ref{fig:bistrie} only serves for illustration purposes.
When using this construction there is a one-to-one correspondence between the nodes and the substrings of lexicon words. In general, the number of substrings of entries in ${\mathcal{D}}$ is quadratic in the size (number of symbols) of ${\mathcal{D}}$.
In the next section we shall describe an  index structure that has the same functionality and needs storage space linear in the size (number of symbols) of the lexicon ${\mathcal{D}}$.
\end{remark}



\begin{remark}
The approach that we proposed in this section is closely related to the algorithm of Myers~\cite{M94} for approximate search in strings. The
main difference is that we have a fixed threshold $b$ for the number of errors, whereas in \cite{M94}
the threshold is given in terms of percentage of symbols.
This imposes different ways of handling the arising situation
and modifications related with the application of the pigeonhole principle. Thus for query words of
length at least $b+1$ we shall always have an initialization with an exact match, whereas in Myers' situation
this assumption is not obligatory fulfilled and he is not able to use it.
\end{remark}

\section{Symmetric compact directed acyclic word graphs}\label{sec:suffix_aut}

In this section we describe the bidirectional index structure for search in the lexicon.
Afterwards we explain how the index structure supports the computation of solutions sets
described in Section~\ref{subsectionComputationSolutionsSets} during online steps.
As before, $\mathcal{D}$ denotes the given lexicon,
$\#\mathcal{D}\$$ denotes the variant where the new symbols $\#$ and $\$$ are attached as the first and the last symbol to each lexicon word,
$\Sigma_{\#\$}= \Sigma \cup \{\#, \$\}$.


\subsection{The index structure}
In Section~\ref{SubSecSuffixTries} we described how suffix tries and suffix tries for reversed words of the lexicon can be merged into a bidirectional
(quadratic) index structure, using a bijective correspondence between the states of the two substructures.
We now introduce a bidirectional index of linear size, which is used in our method for approximate search.
Furthermore we present a new algorithm for online construction of such index.
To build the index we crucially use an algorithm from \cite{Inenaga05} for online construction of one-directional compact directed acyclic word graphs.
\begin{definition}\label{DefStartPosition}
Let $W \in \Sigma_{\#\$}^{\ast}$, let $0\leq i\leq \vert W\vert$. A word $V \in \Sigma_{\#\$}^{\ast}$ of length $0 < \vert V\vert \leq \vert W\vert$
is said to {\em start at position $i$ in $W$} if $i + \vert V\vert - 1 \leq \vert W\vert$ and $W_{i} W_{i + 1} \ldots W_{i + \vert V\vert - 1} = V$. 
Similarly $V$ is said to {\em end at position $i$ in $W$} if $0 < i - \vert V\vert + 1$ and
$W_{i - \vert V\vert + 1} W_{i - \vert V\vert + 2} \ldots W_{i} = V\}$. 
We define the functions $startpos_W$ and $endpos_W$ as
\begin{eqnarray*}
startpos_W(V) &:=& \{i \in \mathbb{N} \mid \mbox{$V$ starts in $W$ at position $i$}\},\\
endpos_W(V) &:=& \{i \in \mathbb{N} \mid \mbox{$V$ ends in $W$ at position $i$}\}.
\end{eqnarray*}
In addition, let
$startpos_W(\varepsilon) := endpos_W(\varepsilon) := \{ 0, 1, \ldots, \vert W \vert \}$.

\noindent We define the equivalence relations $\rightEquiv_W$ and $\leftEquiv_W$ on $\Sigma_{\#\$}^{\ast}$ as:
\begin{eqnarray*}
X \rightEquiv_W Y &\iff& startpos_W(X) = startpos_W(Y),\\
X \leftEquiv_W Y  &\iff& endpos_W(X) = endpos_W(Y).
\end{eqnarray*}
\end{definition}
\begin{definition}
The equivalence relations $\rightEquiv_{\#\mathcal{D}\$}$ and
$\leftEquiv_{\#\mathcal{D}\$}$
are defined on $\Inf(\#\mathcal{D}\$)$ as follows.
For every $X, Y \in \Inf(\#\mathcal{D}\$)$
$$X \rightEquiv_{\#\mathcal{D}\$} Y \iff \forall W \in \#\mathcal{D}\$ : X \rightEquiv_W Y,$$
$$X \leftEquiv_{\#\mathcal{D}\$} Y \iff \forall W \in \#\mathcal{D}\$ : X \leftEquiv_W Y.$$
\end{definition}
In what follows, the equivalence class of a substring
$V \in \Inf(\#\mathcal{D}\$)$ w.r.t. $\rightEquiv_{\#\mathcal{D}\$}$
($\leftEquiv_{\#\mathcal{D}\$}$) is written $\myright{[V]}$ ($\myleft{[V]}$).
It is easy to prove the following properties of the function $startpos_W$ ($endpos_W$).
\begin{proposition}\label{PropositionEndPos1}
Let $W, X, Y \in \Sigma_{\$}^{\ast}$ be arbitrary strings.
\begin{enumerate}
\item\label{PropositionEndPos1Item1}
If $startpos_W(X) \cap startpos_W(Y)\neq \emptyset$ ($endpos_W(X) \cap endpos_W(Y)\neq \emptyset$),
then $X$ is a prefix (suffix) of $Y$ or vice versa,
\item\label{PropositionEndPos1Item2}
If $Y$ is a prefix (suffix) of $X$, then $startpos_W(X) \subseteq startpos_W(Y)$ ($endpos_W(X) \subseteq endpos_W(Y)$).
\end{enumerate}
\end{proposition}
Proposition~\ref{PropositionEndPos1} can be used to show that for any two elements $X, Y \in \myright{[V]}$ ($X, Y \in \myleft{[V]}$)
either $X$ is a prefix (suffix) of $Y$ or vice versa.
Consequently we can define \emph{the canonical representative}
$\myright{X}$ of $\myright{[X]}$
($\myleft{X}$ of $\myleft{[X]}$) as the \emph{longest word}
$\myright{X} \in \myright{[X]}$ ($\myleft{X} \in \myleft{[X]}$).
\begin{proposition}\label{propUnique} {\cite{Inenaga05}}
For every $X \in \Inf(\#\mathcal{D}\$)$ there uniquely exist $\alpha, \beta \in \Sigma_{\#\$}^{\ast}$ such that $\myleft{X} = \alpha X$ and $\myright{X} = X \beta$.
\end{proposition}
\begin{definition} For every $X \in \Inf(\#\mathcal{D}\$)$ we define $\myleftright{X} := \alpha X \beta$ where
$\myleft{X} = \alpha X$ and $\myright{X} = X \beta$.
The equivalence relation $\leftrightEquiv_{\#\mathcal{D}\$}$ is defined on $\Inf(\#\mathcal{D}\$)$ as follows.
For every $X, Y \in \Inf(\#\mathcal{D}\$)$
$$X \leftrightEquiv_{\#\mathcal{D}\$} Y \iff \myleftright{X} = \myleftright{Y}.$$
\end{definition}
In what follows the equivalence class of $X$ w.r.t. $\leftrightEquiv_{\#\mathcal{D}\$}$ is written $\myleftright{[X]}$.
We shall use $\leftEquiv$, $\rightEquiv$ and $\leftrightEquiv$ as shorthands for 
$\leftEquiv_{\#\mathcal{D}\$}$, $\rightEquiv_{\#\mathcal{D}\$}$ and $\leftrightEquiv_{\#\mathcal{D}\$}$ respectively.
\begin{proposition} {\cite{BlumerBlumer87}} The equivalence relation
$\leftrightEquiv$ is the transitive closure of
$\rightEquiv$ and $\leftEquiv$.
\end{proposition}
\begin{proposition}
The equivalence relation $\leftEquiv$ is right-invariant: for arbitrary substrings $X, Y \in \Inf(\#\mathcal{D}\$)$ and
arbitrary extensions of the form $X \circ U, Y \circ U \in \Inf(\#\mathcal{D}\$)$ always
$X \leftEquiv Y$ implies $X \circ U \leftEquiv Y \circ U$.
The equivalence relation $\rightEquiv$ is left-invariant.
\end{proposition}
\begin{definition} {\cite{BlumerBlumer87}} \label{DefinitionDAWG}
The {\em directed acyclic word graph} (DAWG) for $\#\mathcal{D}\$$ is the deterministic finite-state automaton
$\leftA(\#\mathcal{D}\$) := (Q_{\leftA},\Sigma_{\#\$},\myleft{[\varepsilon]},\delta_{\leftA},F_{\leftA})$ where
\begin{itemize}
\item
$Q_{\leftA}$ is the set of all equivalence classes $\myleft{[V]}$  w.r.t. $\leftEquiv$,
\item
the start state is $\myleft{[\varepsilon]}\in Q_{\leftA}$,
\item
the (partial) transition function $\delta_{\leftA}$ is defined as
$\delta_{\leftA}(\myleft{[V]},\sigma)=\myleft{[V \circ \sigma]}$ for all substrings $V \circ \sigma$ of $\#\mathcal{D}\$$,
\item
the set of final states is
$F_{\leftA} := \{ \myleft{[V]} \mid \myleft{[V]} \cap \#\mathcal{D}\$ \neq \emptyset \}$.
\end{itemize}
\end{definition}
Note that the right-invariance of $\leftEquiv$ implies that $\leftA$ is well-defined.
The DAWG for $\#\mathcal{D}\$$ can be used (i) to check if a string $V$ is in $\Inf(\#\mathcal{D}\$)$
and in the positive case (ii) to check in constant time if a right extension $V \circ \sigma$ again represents such a substring:
for solving problem (i) we start a traversal of
$\leftA(\#\mathcal{D}\$)$ from $\myleft{[\varepsilon]}$ with the letters of $V$. Then
$V \in \Inf(\#\mathcal{D}\$)$ iff all transitions are defined. In the positive case the traversal leads to the state $\myleft{[V]}$.
To solve problem (ii) we check if $\delta_{\leftA}(\myleft{[V]},\sigma)$ is defined.
Note that for tasks (i) and (ii) we need not fix a set of final states. With the above definition of final states we may check if
a substring of the form $\#V\$$ represents a full entry of the lexicon. This holds iff $\delta_{\leftA}^{\ast}(\myleft{[\varepsilon]},\#V\$)$ is (defined and) final.
Analogously, using $\rightEquiv$, we define $\rightA(\#\mathcal{D}\$)$, which can be used to check for left extensions $\sigma \circ V$.
The question is how to merge $\leftA$ and $\rightA$ into one bidirectional index.
\begin{definition} {\cite{BlumerBlumer87}} \label{DefinitionCDAWG}
The {\em compact directed acyclic word graph} (CDAWG) for $\#\mathcal{D}\$$ is the generalized deterministic finite-state automaton
$\leftC(\#\mathcal{D}\$) := (Q_{\leftrightC},\Sigma_{\#\$},\myleftright{[\varepsilon]},\delta_{\leftC},F_{\leftrightC})$ where
\begin{itemize}
\item
$Q_{\leftrightC}$ is the set of all equivalence classes $\myleftright{[V]}$  w.r.t. $\leftrightEquiv$,
\item
the start state is $\myleftright{[\varepsilon]}\in Q_{\leftrightC}$,
\item
the (partial) transition function $\delta_{\leftC} : Q_{\leftrightC} \times \Sigma_{\#\$}^{+} \rightarrow Q_{\leftrightC}$ is defined for
 a string of the form $\sigma U$ ($\sigma\in \Sigma_{\#\$}$, $U \in \Sigma_{\#\$}^{\ast}$) iff 
$\myright{\myleftright{V} \circ \sigma} = \myleftright{V} \circ \sigma U$. The value is 
$$\delta_{\leftC}(\myleftright{[V]},\sigma U) = \myleftright{[\myleftright{V} \circ \sigma]}$$
\item
the set of final states is
$F_{\leftrightC} := \{ \myleftright{[V]} \mid \myleftright{[V]} \cap \#\mathcal{D}\$ \neq \emptyset \}$.
\end{itemize}
\end{definition}
$\leftC(\#\mathcal{D}\$)$ can be considered as a compacted variant of $\leftA(\#\mathcal{D}\$)$,
because $\leftC(\#\mathcal{D}\$)$ can be obtained from $\leftA(\#\mathcal{D}\$)$ by replacing
chains of the type $q_{0} \stackrel{\sigma_{1}}{\rightarrow} q_{1} \stackrel{\sigma_{2}}{\rightarrow} q_{2} \ldots \stackrel{\sigma_{n}}{\rightarrow} q_{n}$
with multi-letter transitions
$q_{0} \stackrel{\sigma_{1}\sigma_{2} \ldots q_{n-1} \sigma_{n}}{\longrightarrow} q_{n}$ iff
states $q_{i}$ for $1 < i < n$ are implicit\footnote{A state $q$ is called {\em implicit} iff
$q$ is not the start state, $q$ is not final and $q$ has exactly one outgoing transition. A state is called {\em explicit} iff it is not implicit.}
and $q_{1}$ and $q_{n}$ are explicit, \cite{BlumerBlumer87}. Analogously we define $\rightC(\#\mathcal{D}\$)$ with 
(partial) transition function $\delta_{\rightC}$:
$$\delta_{\rightC}(\myleftright{[V]},\sigma U) = \myleftright{[\sigma \circ \myleftright{V}]} \iff
\myleft{\sigma \circ \myleftright{V}} = U^{rev} \sigma \circ \myleftright{V}.$$
$\rightC(\#\mathcal{D}\$)$ can be considered as well as a compacted variant of $\rightA(\#\mathcal{D}\$)$ in the above sense.
Note that both automata have the same set of states.
Hence $\leftC(\#\mathcal{D}\$)$ and $\rightC(\#\mathcal{D}\$)$ are naturally merged into one bidirectional index. The following index is used in our method for approximate search.
\begin{definition} {\cite{Inenaga01}} \label{DefinitionSCDAWG}
The bidirectional {\em symmetric compact acyclic word graph} (SCDAWG) for $\#\mathcal{D}\$$ is 
$$\leftrightC(\#\mathcal{D}\$) := (Q_{\leftrightC},\Sigma_{\#\$}^{+},\myleftright{[\varepsilon]},\delta_{\leftC}, \delta_{\rightC}, F_{\leftrightC}).$$
\end{definition}

\noindent{\bf Linear description of SCDAWGs.\ } Our next goal is to show how to represent $\leftrightC(\#\mathcal{D}\$)$ in space linear in the lexicon. 
\begin{proposition}\label{propLinearNumbersSCDAWG} 
The following inequalities hold for the number of states $\vert Q_{\leftrightC}\vert$,
the number of the transitions in $\delta_{\leftC}$, $\vert \delta_{\leftC}\vert$, and
the number of the transitions in $\delta_{\rightC}$, $\vert \delta_{\rightC}\vert$, in the SCDAWG $\leftrightC(\#\mathcal{D}\$)$:
$$\vert Q_{\leftrightC}\vert \leq 2 \vert\vert \#\mathcal{D} \vert\vert,$$
$$max( \vert \delta_{\leftC}\vert, \vert \delta_{\rightC}\vert ) \leq 2 \vert\vert \#\mathcal{D} \vert\vert - 1.$$
\end{proposition}
{\em Proof.\ }
First we shall give upper bounds for size of the {\em suffix tree} for $\#\mathcal{D}\$$,
defined as the generalized deterministic finite-state automaton
$STree(\#\mathcal{D}\$) := (Q, \Sigma_{\#\$}, \myright{\varepsilon}, \delta, F)$ where
$Q := \{ \myright{X} \vert X \in \Inf(\#\mathcal{D}\$) \}$, $F := \Suff(\#\mathcal{D}\$)$ and
for $\sigma \in \Sigma_{\#\$}$ and $U \in \Sigma_{\#\$}^{+}$,
$\delta( \myright{X}, \sigma U ) = \myright{\myright{X} \circ \sigma}$ iff $\myright{X} \circ \sigma U \in Q$ and
$\myright{\myright{X} \circ \sigma} = \myright{X} \circ \sigma U$, \cite{Inenaga01}.
The suffix tree represents a tree with root $\myright{\varepsilon}$ and leaves $F$.
For the number of the leaves we have $\vert F \vert \leq 1 + \vert\vert \#\mathcal{D} \vert\vert$.
Each internal node $\myright{X}$ of the suffix tree has at least two successors.
Hence $\vert Q\vert \leq 2 \vert\vert \#\mathcal{D} \vert\vert$.
For the number of transitions in the suffix tree we have $\vert \delta\vert \leq 2 \vert\vert \#\mathcal{D} \vert\vert - 1$.
For every $X \in \Inf(\#\mathcal{D}\$)$ it can be shown that $\myleftright{X} \in Q$ and
for every transition $\myright{[X]} \stackrel{\sigma U}{\rightarrow} \myleftright{[Y]}$ in $\delta_{\leftC}$
it can be shown that there is a suffix tree transition $\myleftright{X} \stackrel{\sigma U}{\rightarrow} \myleftright{X} \sigma U$ in $\delta$.
Consequently $\vert Q_{\leftrightC}\vert \leq \vert Q\vert$ and $\vert \delta_{\leftC}\vert \leq \vert \delta\vert$.
Analogously we obtain that $\vert \delta_{\rightC}\vert$ is bounded by the number of transitions in $STree(\$\mathcal{D}^{rev}\#)$.
\qed
\begin{remark}\label{remarkLinarRepresentation} Note that Proposition~\ref{propLinearNumbersSCDAWG} is not sufficient to prove that
the size of SCDAWG is $O(\vert \vert \#\mathcal{D}\$ \vert \vert)$, since the labels of the transitions are strings in $\Sigma_{\#\$}^{+}$.
To achieve a linear description of SCDAWG the transitions are represented as follows.
Let $D$ be a concatenation of all strings in $\#\mathcal{D}\$$. For every state $q = \myleftright{[Y]}$
we store a position $end(q) = i$ in $D$ where $\myleftright{Y}$ terminates,
$\myleftright{Y} = D_{i - \vert \myleftright{Y}\vert + 1} D_{i - \vert \myleftright{Y}\vert + 2} \ldots D_{i}$.
For every transition $t = \myleftright{[X]} \stackrel{\alpha}{\rightarrow} \myleftright{[Y]}$ in $\delta_{\leftC}$
let $start(t) := end(\myleftright{[Y]}) - \vert \alpha\vert + 1$.
Since $\myleftright{X} \alpha$ is suffix of $\myleftright{Y}$, for every transition $t$ in $\delta_{\leftC}$ we store only $start(t)$,
but not the whole label $\alpha$. In analogous way we define $start(q)$ and $end(t)$ and for every transition $t$ in $\delta_{\rightC}$
we store only $end(t)$.
\end{remark}

\noindent{\bf Online construction of SCDAWGs in linear time.\ }
In \cite{Inenaga01} Inenaga et al. present an online algorithm that builds letter by letter the SCDAWG $\leftrightC(\{\#W\$\})$ for a single string $\#W\$$
in time $O(\vert \#W\$\vert)$.
Here we present a new straightforward online algorithm that builds
a representation of $\leftrightC(\#\mathcal{D}\$)$ string by string in time $O(\vert\vert\#\mathcal{D}\$\vert\vert)$.
Our result is essentially based on another algorithm by Inenaga et al. \cite{Inenaga05},
that constructs $\leftC(\vert \#\mathcal{D}\$\vert)$ letter by letter in an online manner.
The idea is to synchronize $\leftC(\#\mathcal{D}\$)$ and $\leftC(\$\mathcal{D}^{rev}\#)$ while simultaneously building both of them word by word.
\begin{proposition} $\rightC(\#\mathcal{D}\$)$ and $\leftC(\$\mathcal{D}^{rev}\#)$ are isomorphic.
\end{proposition}
{\em Proof.\ } 
Let
$\rightC(\#\mathcal{D}\$) = (Q,\Sigma_{\#\$},\myleftright{[\varepsilon]},\delta_{\rightC},F)$ and
$\leftC(\$\mathcal{D}^{rev}\#) = (Q',\Sigma_{\#\$},\myleftright{[\varepsilon]},\delta',F')$.
The isomorphism is given by the bijection $b : Q \rightarrow Q'$ defined as follows.
$$b(\myleftright{[X]}) := \myleftright{[X^{rev}]}.$$ 
\qed
\noindent
Since $\leftC(\#\mathcal{D}\$)$ and $\rightC(\#\mathcal{D}\$)$ have one and the same set of states,
the bijection $b$, defined in the above proof,
provides the way to express $\delta_{\rightC}$ as follows:
$$\delta_{\rightC}(q, \alpha) = b^{-1}(\delta'(b(q), \alpha)).$$
In our online construction we compute $\delta_{\leftC}$, $\delta'$ and all values of $b$ and $b^{-1}$ for every state $q$.
Let us note that if we directly compute $b(\myleftright{[X]})$ for a given state $\myleftright{[X]} \in Q$ by reversing some $Y \in \myleftright{[X]}$
and traversing $\leftC(\$\mathcal{D}^{rev}\#)$ with $Y^{rev}$ from the initial state,
the total time for the whole construction would be in the worst case quadratic w.r.t. $\vert\vert \#\mathcal{D}\$\vert\vert$.
To achieve linear time we need to compute $b(\myleftright{[X]})$, given a state $\myleftright{[X]}$, in amortized time $O(1)$.
We show that such an efficient online computation of $b$ can be based on the {\em suffix links} provided for every state $\myleftright{[X]}$
during the construction of $\leftC(\#\mathcal{D}\$)$.
\begin{definition} The {\em suffix link} $sl(\myleftright{[X]})$ of a state $\myleftright{[X]}$ in $\leftC(\#\mathcal{D}\$)$
is $\myleftright{[Y]}$ where $\myleftright{Y}$ is the longest suffix of $\myleftright{X}$ such that $\myleftright{Y} \not\in \myleftright{[X]}$.
$sl(\myleftright{[\varepsilon]})$ is not defined.
\end{definition}
\begin{proposition}\label{propSuffixLinks} Let 
\begin{eqnarray*}
\leftC(\#\mathcal{D}\$) &=& (Q,\Sigma_{\#\$}, \myleftright{[\varepsilon]}, \delta_{\leftC}, F),\\
\leftC(\$\mathcal{D}^{rev}\#) &=& (Q',\Sigma_{\#\$},\myleftright{[\varepsilon]},\delta',F'),
\end{eqnarray*}
$\myleftright{[X]} \in Q$, $sl(\myleftright{[X]}) = \myleftright{[Y]}$ and $b(\myleftright{[Y]}) = q'$.
Let $j = \vert \myleftright{X}\vert - \vert\myleftright{Y}\vert$, $\sigma = \myleftright{X}_{j}$. Then
there exists a $\sigma$-transition from $q'$. Let $\delta'(q', \sigma U) = p'$ be the $\sigma$-transition from $q'$.
Then $b(\myleftright{[X]}) = p'$.
\end{proposition}
{\em Proof.\ }
The $\sigma$-transition from $q' = \myleftright{[Y^{rev}]}$ is defined,
since $\myleftright{Y^{rev}}$ is prefix of $\myleftright{X^{rev}}$, which implies that there is a path
with label $\myleftright{X}_{j} \myleftright{X}_{j-1} \ldots \myleftright{X}_{1}$
in $\leftC(\$\mathcal{D}^{rev}\#)$ from $q'$ to $\myleftright{[X^{rev}]}$.
If we assume that this path is not composed of one single transition, then for the last intermediate state $\myleftright{[Z^{rev}]}$ of this path
we have that $\myleftright{Z}$ is suffix of $\myleftright{X}$, $\myleftright{Z}$ is longer than $\myleftright{Y}$ and
$\myleftright{Z} \not\in \myleftright{[X]}$,
which contradicts with $sl(\myleftright{[X]}) = \myleftright{[Y]}$.
\qed
\noindent The algorithm of Figure~\ref{fig:ComputeSCDAWG} calculates $\leftC(\#\mathcal{D}\$)$, $\leftC(\$\mathcal{D}^{rev}\#)$, $b$ and $b^{-1}$,
 given the lexicon $\#\mathcal{D}\$$. The states of these two CDAWGs are consecutive integers starting from $0$, which is the initial state.
The function $AddStringInCDAWG( \leftC, \#W\$ )$ represents the online construction of CDAWG invented by Inenaga et al., \cite{Inenaga05}.
$AddStringInCDAWG$ adds the string $\#W\$$ to $\leftC$.
$AddStringInCDAWG$ changes its first argument $\leftC$ by adding new consecutive states in $Q$ and $F$ and by setting the transition function $\delta_{\leftC}$ and
the suffix links $sl$ for every new state. $AddStringInCDAWG$ never changes $\myleftright{X}$ for every state $\myleftright{[X]}$ that is already in $Q$.
Hence the computation of $b$ is stable in the sense that once $b(q)$ is precomputed for a given state $q$, further changes of $b(q)$ are impossible. 
Based on Proposition~\ref{propSuffixLinks} the function $FindState$ recursively calculates the values of $b$.
In line $5$ of $FindState$ we use $end(i)$, the concatenation $D$ of the strings accumulated so far described in Remark~\ref{remarkLinarRepresentation}
and the {\em length} of state $s$ defined as the length of the longest member of the equivalence class represented by $s$.
The lengths of the states are computed by $AddStringInCDAWG$.
The bottom of the recursion is guaranteed by $b(0) = 0$ and the decreasing lengths of the input states provided in recursive calls.
The number of times $FindState$ is invoked is $O(\vert Q\vert)$. Since the time for the online construction of CDAWG is $O(\vert\vert \#\mathcal{D}\$\vert\vert)$
we obtain the following.
\begin{proposition}
The online algorithm on Figure~\ref{fig:ComputeSCDAWG} runs in time $O(\vert\vert \#\mathcal{D}\$\vert\vert)$.
\end{proposition}
\begin{figure}
\noindent
$BuildSDAWG( \: \#\mathcal{D}\$ \: )\{$\\
\verb@1     @\mbox{$Q = \{0\}; \: \delta = \emptyset; \: F = \emptyset; \: Q' = \{0\}; \: \delta' = \emptyset; \: F' = \emptyset;$}\\
\verb@2     @$\leftC = (Q,\Sigma_{\#\$},0,\delta,F); \: \leftC' = (Q',\Sigma_{\#\$},0,\delta',F'); \: b = \emptyset; \: b^{-1} = \emptyset;$\\
\verb@3     @$b(0) = 0; \: b^{-1}(0) = 0;$\\
\verb@4     @\mbox{$\keyword{for}( \: \#W\$ \in \#\mathcal{D}\$ \: )\{$}\\
\verb@5         @$n = \vert Q\vert;$\\
\verb@6         @$AddStringInCDAWG( \leftC, \#W\$ );$\\
\verb@7         @$AddStringInCDAWG( \leftC', \$W^{rev}\# );$\\
\verb@8         @$i = n;$\\
\verb@9         @\mbox{$\keyword{while}( \: i < \vert Q\vert \: )\{$}\\
\verb@10            @$b(i) = nil; \: i = i + 1;$\\
\verb@11        @$\}$\\
\verb@12        @$i = n;$\\
\verb@13        @\mbox{$\keyword{while}( \: i < \vert Q\vert \: )\{$}\\
\verb@14            @$\keyword{if}( \: b(i) == nil \: )\{$\\
\verb@15                @$b(i) = FindState( \leftC, \leftC', b, b^{-1}, i ); \: b^{-1}(b(i)) = i;$\\
\verb@16            @$\}$\\
\verb@17            @$i = i + 1;$\\
\verb@18        @$\}$\\
\verb@19    @$\}$\\
\verb@20    @$\keyword{return} \:\: (\leftC, \leftC', b, b^{-1});$\\
$\}$\\
$FindState( \: \leftC, \leftC', b, b^{-1}, i \: )\{$\\
\verb@1     @$s = sl(i);$\\
\verb@2     @$\keyword{if}( \: b(s) == nil \: )\{$\\
\verb@3         @$b(s) = FindState( \leftC, \leftC', b, b^{-1}, s ); \: b^{-1}(b(s)) = s;$\\
\verb@4     @$\}$\\
\verb@5     @$q' = b(s); j = end(i) - length(s); \sigma = D_{j};$\\
\verb@6     @\mbox{let $\delta'(q', \sigma U) = p'$ be the $\sigma$-transition from $q'$ in $\leftC'$}\\
\verb@7     @$\keyword{return} \:\: p';$\\
$\}$\\
\caption{Online construction of a representation of SCDAWG for $\#\mathcal{D}\$$ as
$(\stackrel{\leftarrow}{C}(\#\mathcal{D}\$), \stackrel{\leftarrow}{C}(\$\mathcal{D}^{rev}\#), b, b^{-1})$}\label{fig:ComputeSCDAWG}
\end{figure}
The SCDAWG for $\#\mathcal{D}\$$ can be considered as a compact version of the
bidirectional suffix trie $BiSTrie(\#\mathcal{D}\$)$, Definition~\ref{DefinitionBiSTrie}.
\begin{example}
The SCDAWG for the example lexicon 
$$\#\mathcal{D}\$ = \{ {\tt \#ear\$}, {\tt \#lead\$}, {\tt \#real\$} \}$$ 
is shown in Figure~\ref{fig:scdawg}.
The dashed transitions represent $\delta_{\rightC}$, while the solid transitions represent $\delta_{\leftC}$.
The equivalence classes are
$0$ - $\{ \varepsilon \}$,
$1$ - $\{\tt l \}$,
$2$ - $\{\tt \# \}$,
$3$ - $\{\tt \$ \}$,
$4$ - $\{\tt a, ea \}$,
$5$ - $\{\tt r \}$,
$6$ - $\{\tt d\$, ad\$, lead\$, \#lead\$ \}$,
$7$ - $\{\tt l\$, al\$, eal\$, real\$, \#real\$ \}$ and
$8$ - $\{\tt r\$, ar\$, ear\$, \#ear\$ \}$.
\begin{figure}
\begin{center}
\includegraphics[width=\textwidth]{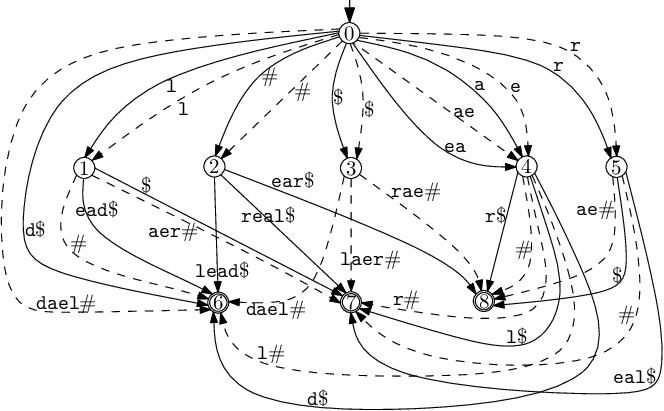}
\caption{SCDAWG for $\{ {\tt \#ear\$}, {\tt \#lead\$}, {\tt \#real\$} \}$.}
\label{fig:scdawg}
\end{center}
\end{figure}
\end{example}

\subsection{Bidirectional online search using SCDWAGs}

We now describe how the above index structure is used for computation of solution sets defined in Section~\ref{subsectionComputationSolutionsSets}.
We assume that the following {\bf offline resources} are available:
\begin{enumerate}
\item\label{item1} the SCDAWG $\leftrightC(\#\mathcal{D}\$)$, in particular the two transition funtions $\delta_{\leftC}$ and $\delta_{\rightC}$;
\item\label{item2} $start(\myleftright{[V]})$ and $end(\myleftright{[V]})$ for every state $\myleftright{[V]}$, Remark~\ref{remarkLinarRepresentation};
\item\label{item3} $start(t)$ for every transition in $\delta_{\leftC}$, Remark~\ref{remarkLinarRepresentation};
\item\label{item4} $end(t)$ for every transition in $\delta_{\rightC}$, Remark~\ref{remarkLinarRepresentation}.
\end{enumerate}
We keep track of the following {\bf online information} - here $W$ denotes the substring of a lexicon word faced at a certain point of the computation of solution sets
and $D$ denotes the concatenation used in the linear representation of the SCDAWG $\leftrightC(\#\mathcal{D}\$)$,~\ref{remarkLinarRepresentation}.
\begin{enumerate}
\item\label{item5} the length of $W$;
\item\label{item6} (the number of) the state $\myleftright{[W]}$;
\item\label{item7} the unique (Proposition~\ref{propUnique}) position $j_W$ of $W$ in $D$ such that $start(\myleftright{[W]}) \leq j_W \leq end(\myleftright{[W]})$.
\end{enumerate}
Let $\sigma \in \Sigma_{\#\$}$.
We consider possible extensions of the current substring $W$ to the right of the form $W \sigma$ as follows.
If $j_W + \vert W\vert \leq end(\myleftright{[W]})$, then $W \sigma \in \Inf(\#\mathcal{D}\$)$ iff $\sigma = D_{j_W + \vert W\vert}$
and if $\sigma = D_{j_W + \vert W\vert}$, then $\myleftright{[W \sigma]} = \myleftright{[W]}$ and $j_{W \sigma} = j_W$.
If $j_W + \vert W\vert > end(\myleftright{[W]})$, then $W \sigma \in \Inf(\#\mathcal{D}\$)$ iff
there exists a $\sigma$-transition from $\myleftright{[W]}$ in $\delta_{\leftC}$.
Let $t = \myleftright{[W]} \stackrel{\sigma U}{\rightarrow} \myleftright{[V]}$ be the $\sigma$-transition from $\myleftright{[W]}$ in $\delta_{\leftC}$.
Then $\myleftright{[W \sigma]} = \myleftright{[V]}$ and $j_{W \sigma} = start(t) - \vert W\vert$.
Possible extensions to the left of the form $\sigma W$ are handled similarly by using $start(\myleftright{[W]})$ and $end(t)$.
\begin{example}\label{FinalExample}
One example for the use of $\leftrightC(\#\mathcal{D}\$)$ in Figure~\ref{fig:scdawg} is the following.
We first want to check if ${\tt e}$ is a substring in $\Inf(\#\mathcal{D}\$)$.
For this aim we start from state $0$ and follow the ${\tt e}$-transition $t = 0 \stackrel{\tt ea}{\rightarrow} 4$ in $\delta_{\leftC}$,
$j_{\tt e} = start(t) = start(4) = end(4) - 1$, the number of the state $\myleftright{[\tt e]}$ is $4$.
\begin{itemize}
\item
We now want to find all left extensions with a single letter.
Since $j_{\tt e} = start(4)$, we have to use the three possible dashed transitions from state $4$.
With $\#$ we reach $8$, the number of the state $\myleftright{[\tt \#e]}$ is $8$,
$j_{\tt \#e} = start(8) = end(8) - 4$.
With ${\tt r}$ we reach $7$, the number of the state $\myleftright{[\tt re]}$ is $7$,
$j_{\tt re} = start(7) + 1 = end(7) - 4$.
With ${\tt l}$ we reach $6$, the number of the state $\myleftright{[\tt le]}$ is $6$,
$j_{\tt le} = start(6) + 1 = end(6) - 4$.
\item
We now want to find all right extensions of ${\tt e}$ with a single letter.
Since $j_{\tt e} + 1 = end(4)$, the only one possible extension to the right is with letter ${\tt a}$,
the number of the state $\myleftright{[\tt ea]}$ is $4$,
$j_{\tt ea} = j_{\tt e} = start(4) = end(4) - 1$.
If we want to further extend ${\tt ea}$
to the right, we have to use the solid transition, because $j_{\tt ea} + 2 > end(4)$.
\end{itemize}
\end{example}


\begin{remark}
In our actual implementation we use a simple optimization of the approximate search based on additional information stored in the SCDAWG. The idea is to use ``positional'' information to recognize blind paths of the search. 
Consider a substring $V$ of a lexicon word. If some lexicon word 
$W$ has the form $W=RVS$ we say that $\vert R\vert$ (resp. $\vert S\vert$) is the {\em length of a possible prefix (suffix) for $V$}.
In the SCDAWG we store for each substring 
$\myleftright{V}$ the maximal and minimal length of a possible prefix (suffix) for $\myleftright{V}$.
When computing solution sets $Sol_{\mathcal{D}}(P', b')$, substrings $P_0$ of the pattern $P$ are aligned with substrings $V$ found in the SCDAWG. Each substring $P_0$ defines a unique prefix $R_P$ and a unique suffix $S_P$ of the pattern $P=R_PP_0S_P$. 
We check if the length of $R_P$ and $S_P$ is ``compatible'' with the information stored in the SCDAWG for the length of possible prefixes and suffixes for $V$. To test ``compatibility'', the error bound and the distance between $P_0$ and $V$ has to be taken into account. Compatibility is checked each time we reach new  state of the SCDAWG.
We omit the technical details.
\end{remark}

\section{Evaluation}\label{sec:evaluation}

In this section we compare our new method to two other methods for efficient approximate search in lexica, Oflazer's approach \cite{Kemal} and the forward-backward method introduced in \cite{MS04}.
To have a common basis for the experiments we always use as a filter mechanism Ukonnen's optimized matrix method \cite{Uk85}.\footnote{Universal Levenshtein automata \cite{SM02,MS04} are more efficient but can only be built for small distance bounds because of huge memory requirements.}
For the three methods we present experimental results for approximate search in lexica of different sizes and types. We also look at the dependency of search times on the notion of similarity used.
In order to get a picture of principle limitations for approximate search we also present evaluation results where we simulate the ``ideal method''.

The ``ideal method'' for bound $b \in \mathbb{N}$ and dictionary ${\mathcal{D}}$
is based on a {\em perfect index} $I({\mathcal{D}},b)$ that directly maps every query $(P, b)$ to
the solution $Sol_{\mathcal{D}}(P, b) \cap \mathcal{D}$. Since the size of the perfect index $I({\mathcal{D}},b)$ would be too large,
for every experiment we build a {\em restricted perfect index} $I(T,{\mathcal{D}},b)$
that works only for a small finite test set ${\cal T} \subset \Sigma^\ast$ of query strings.
For every query string $P \in {\cal T}$ the restricted perfect index $I({\cal T},{\mathcal{D}},b)$ maps $(P, b)$ to the solution
$Sol_{\mathcal{D}}(P, b) \cap \mathcal{D}$.
We represent the restricted perfect index $I({\cal T},{\mathcal{D}},b)$ as an acyclic $k$-subsequential transducer for
$k = max_{P \in {\cal T}} \vert Sol_{\mathcal{D}}(P, b) \cap \mathcal{D}\vert$.
An online algorithm for building minimal acyclic $k$-subsequential transducers is introduced in \cite{MM01}.
This form of representation is optimal since the only time used is the time for reading the input and directly producing the desired output.

\subsection{Comparison of search times for different methods} 

For our first series of experiments we chose a lexicon ${\mathcal{D}}$ of $1,200,070$ book titles.
The average length of titles is $47.64$. The number of different symbols in the alphabet of the lexicon is $99$.
We compare search times obtained for
 Oflazer's method \cite{Kemal}, the forward-backward method \cite{MS04}, the new method and the ``ideal method''.
In all experiments we set the weight of each nonidentity edit operation $\op$ to $w(\op) = 1$.
We then vary the distance bound $b$ from $2$ to $15$.
For each bound $b$ we generated a test set ${\cal T}$ of $10,000$ query strings.
Each query string $P$ was received from a randomly chosen string $W \in {\mathcal{D}}$ by
applying randomly $b$ operations from the set of edit operations $\Op$ to $W$ such that $\vert P \vert \geq 3 b$.

All experiments were run on a machine with $64$ gygabytes of RAM, two $2.4$ GHz Quad-Core Intel Xeon $8$-core processors,
$256$ KB L2 cache memory per core and $12$ MB L3 cache memory per processor.
Our implementation uses only one thread.
The amount of memory needed for our experiments is determined by the size of the precomputed index\footnote{We use depth first implementations of the evaluated methods, see Remark~\ref{remarkDFSTraversalIsPossible}.}.

Table~\ref{tableBasic} presents results obtained for the standard Levenshtein distance.
Column 1 specifies the value of the distance bound $b$ used in the experiments.
Explicit search times are only presented for the ideal method (column $5$, times in milliseconds).
Numbers  $x$ in Table~\ref{tableBasic} for some method $M$ mean that the ideal method was $x$ times faster than method $M$ for the problem class.
For example, the entry $13.87$ found in row/column $2$ indicates that approximate search using the new method
presented above with distance bound $2$ and standard Levenshtein distance on average took $13.87$ times the time needed by the ideal method.
Here, as in all experiments, the time needed to write the output words is always included.
Empty cells found in the table mean that we did not wait for the respective method to finish.

\begin{table}
{\scriptsize
\noindent\makebox[\textwidth]{
\begin{tabular}{|c|c|c|c|c|}\hline
\multicolumn{5}{|c|}{Levenshtein distance, lexicon of book titles}         \\ \hline
b	& new / ideal &  fb / ideal & f / ideal  & ideal (ms)\\ \hline
 $2$                 &  $13.87$    &  $325.083$  &  $3817.47$ & $0.004$ \\ \hline
 $3$                 &  $22.80$    &  $686.037$  & $20063.82$ & $0.004$ \\ \hline
 $4$                 &  $50.72$    & $3904.30$   &            & $0.004$ \\ \hline
 $5$                 &  $54.24$    & $7741.55$   &            & $0.005$ \\ \hline
 $6$                 &  $76.36$    &             &            & $0.005$ \\ \hline
 $7$                 &  $86.55$    &             &            & $0.005$ \\ \hline
 $8$                 & $173.98$    &             &            & $0.006$ \\ \hline
 $9$                 & $154.32$    &             &            & $0.006$ \\ \hline
$10$                 & $172.49$    &             &            & $0.006$ \\ \hline
$11$                 & $163.06$    &             &            & $0.007$ \\ \hline
$12$                 & $287.94$    &             &            & $0.007$ \\ \hline
$13$                 & $261.13$    &             &            & $0.007$ \\ \hline
$14$                 & $301.03$    &             &            & $0.008$ \\ \hline
$15$                 & $300.05$    &             &            & $0.008$ \\ \hline
\end{tabular}}
}
\caption{Comparison of search times for four different methods, standard Levenshtein distance, dictionary of titles.
f - Oflazer's method, fb - the forward-backward method, new - the new method, ideal - the ``ideal method''.
Empty cells mean that we did not wait for termination. Explicit search times (in ms) are only given for the ideal method (last column). All other entries represent factors, comparing the given method with the ideal method.}\label{tableBasic}
\end{table}

The results in Table~\ref{tableBasic} show that the method presented in this paper comes ``close'' to the ideal method for small distance bounds when using the standard Levenshtein distance. For the given lexicon of titles, which contains long strings, the new method is dramatically faster than the forward-backward method, which in turn is much faster than Oflazer's method. It is worth to note that the differences become more and more drastic when using larger distance bounds. For these bounds only the new method leads to acceptable search times. 

\subsection{Comparison of search times for language databases with sentences}

For our second series of experiments we use a collection of sentences from the life sciences and biomedical domain.
The lexicon consists of all sentences from $43,000$ paragraphs which were randomly chosen from MEDLINE abstracts\footnote{MEDLINE is a bibliographic database of U.S. National Library of Medicine. MEDLINE contains over 19 million references to journal articles in life sciences, www.nlm.nih.gov/pubs/factsheets/medline.html.}.
The number of sentences in our list is $351,008$. The average number of symbols per sentence is $149.26$.
The size of the lexicon is approximately the same as the size of the lexicon of titles, but the strings are longer.
Table~\ref{tableMEDLINE} presents the comparison of the different methods for the standard Levenshtein distance.
As a new challenge, the distance bound $b$ used for approximate search varies from $2$ to $50$. Note that for previous methods the use of larger distance bounds leads to unacceptable search times. 
Speed-up factors are similar to those observed in Table~\ref{tableBasic}.

\begin{table}
{\scriptsize
\noindent\makebox[\textwidth]{
\begin{tabular}{|c|c|c|c|}\hline
\multirow{2}{*}{$b$} & \multicolumn{3}{c|}{Levenshtein distance, lexicon of MEDLINE sentences}        \\ \cline{2-4}
                     & new / ideal  &  fb / ideal & f / ideal  \\ \hline
 $2$                 &     $8.64$   &  $71.62$    &  $664.91$  \\ \hline
 $3$                 &    $10.72$   &  $145.60$   & $2670.98$  \\ \hline
 $4$                 &    $13.55$   &  $727.22$   &            \\ \hline
 $5$                 &    $16.57$   & $1357.87$   &            \\ \hline
 $6$                 &    $24.54$   &             &            \\ \hline
 $7$                 &    $27.23$   &             &            \\ \hline
 $8$                 &    $36.47$   &             &            \\ \hline
 $9$                 &    $40.68$   &             &            \\ \hline
$10$                 &    $59.48$   &             &            \\ \hline
$11$                 &    $62.36$   &             &            \\ \hline
$12$                 &   $123.31$   &             &            \\ \hline
$13$                 &   $123.84$   &             &            \\ \hline
$14$                 &   $145.98$   &             &            \\ \hline
$15$                 &   $146.11$   &             &            \\ \hline
$20$                 &   $411.71$   &             &            \\ \hline
$30$                 &  $1552.28$   &             &            \\ \hline
$40$                 &  $4872.49$   &             &            \\ \hline
$50$                 & $23869.91$   &             &            \\ \hline
\end{tabular}}
}
\caption{Comparison of search times for different methods, standard Levenshtein distance, dictionary of MEDLINE sentences.
f - Oflazer's method, fb - the forward-backward method, new - the new method, ideal - the ``ideal method''. All entries represent factors, comparing the given method with the ideal method.
Empty cells mean that we did not wait for termination.}\label{tableMEDLINE}
\end{table}

\subsection{Comparison of search times for different variants of Levenshtein distance} 

For our third series of experiments we
compare search times obtained for three notions of similarity, (i) the standard Levenshtein distance,
(ii) the variant where transpositions of neighbored symbols are treated as additional edit operations,
and (iii) the variant where also merges and splits are used as additional edit operations.
Tables~\ref{tableTitles1} (resp. Table~\ref{tableTitles2}) presents results obtained for the variant of Levenshtein distance where we also use
transpositions (merges and splits) as operations.

\begin{table}
{\scriptsize
\noindent\makebox[\textwidth]{
\begin{tabular}{|c|c|c|c|}\hline
\multicolumn{4}{|c|}{Levenshtein distance with transpositions, lexicon of book titles}         \\ \hline
b	& new / ideal &  fb / ideal & f / ideal  \\ \hline
 $2$                 &  $19.69$    &  $346.28$   &  $4081.83$ \\ \hline
 $3$                 &  $36.00$    & $1022.59$   & $20536.91$ \\ \hline
 $4$                 &  $93.36$    & $4767.08$   &            \\ \hline
 $5$                 & $108.54$    & $11861.07$  &            \\ \hline
 $6$                 & $149.84$    &             &            \\ \hline
 $7$                 & $167.16$    &             &            \\ \hline
 $8$                 & $315.19$    &             &            \\ \hline
 $9$                 & $271.12$    &             &            \\ \hline
$10$                 & $324.32$    &             &            \\ \hline
$11$                 & $310.59$    &             &            \\ \hline
$12$                 & $467.27$    &             &            \\ \hline
$13$                 & $459.23$    &             &            \\ \hline
$14$                 & $500.79$    &             &            \\ \hline
$15$                 & $496.84$    &             &            \\ \hline
\end{tabular}
}
}
\caption{Comparison of search times for the variant of Levenshtein distance transpositions of neighbored symbols are treated as edit operations. 
f - Oflazer's method, fb - the forward-backward method, new - the new method, ideal - the ``ideal method''. All entries represent factors, comparing the given method with the ideal method.
Empty cells mean that we did not wait for termination.}\label{tableTitles1}
\end{table}

\begin{table}
{\scriptsize
\noindent\makebox[\textwidth]{
\begin{tabular}{|c|c|c|c|}\hline
\multicolumn{4}{|c|}{Levenshtein distance with merges and splits, lexicon of book titles}         \\ \hline
b	& new / ideal &  fb / ideal & f / ideal  \\ \hline
2	& $51.69$    &  $2297.07$  & $37428.91$ \\ \hline
3	& $155.55$    &  $7199.24$  &            \\ \hline
4	& $693.20$    & $42141.64$  &            \\ \hline
5	& $788.27$    & $112676.78$ &            \\ \hline
6	& $1091.35$   &             &            \\ \hline
7	& $1229.88$   &             &            \\ \hline
8	& $2267.31$   &             &            \\ \hline
9	& $1868.38$   &             &            \\ \hline
10	& $2462.66$   &             &            \\ \hline
11	& $2151.49$   &             &            \\ \hline
12	& $3352.05$   &             &            \\ \hline
13	& $3077.23$   &             &            \\ \hline
14	& $3350.26$   &             &            \\ \hline
15	& $3210.70$   &             &            \\ \hline
\end{tabular}
}
}
\caption{Comparison of search times for the variant of Levenshtein distance where merges of two symbols into one and splits of a symbol into two symbols are treated as edit operations. 
f - Oflazer's method, fb - the forward-backward method, new - the new method, ideal - the ``ideal method''. All entries represent factors, comparing the given method with the ideal method.
Empty cells mean that we did not wait for termination.}\label{tableTitles2}
\end{table}

Basically, the results in Tables~\ref{tableTitles1} and~\ref{tableTitles2} are similar to the previous results: also for the modified distances, the new method is much faster than previous methods, with a speed-up factor of at least 15.  For large distance bounds the speed-up factor is larger.

\subsection{Comparison of search times for different symbol distributions} 

In our fourth series of experiments we ask how the statistical properties of the distribution of letters in the lexicon words influence search times.
We generated two random dictionaries of $1,200,070$ strings -
one with uniform distribution of $99$ symbols and average string length $54.35$
and another one with binomial distribution of $99$ symbols and average string length $54.63$.
In Table~\ref{tableUniform} the columns ``Lev+Binomial'' and ``Lev+Uniform'' present
the behavior of the algorithms for the two random dictionaries
with binomial and uniform distributions of the symbols.

\begin{table}
{\scriptsize
\noindent\makebox[\textwidth]{
\begin{tabular}{|c|c|c|c|}\hline
\multirow{2}{*}{$b$} & \multicolumn{3}{c|}{Lev+Binomial}       \\ \cline{2-4}
                     & new / ideal &  fb / ideal & f / ideal   \\ \hline
 $2$                 &  $18.12$    &   $435.29$  &  $12513.00$ \\ \hline
 $3$                 &  $20.12$    &   $861.60$  & $103772.38$ \\ \hline
 $4$                 &  $22.04$    & $11936.27$  &             \\ \hline
 $5$                 &  $25.30$    & $23053.83$  &             \\ \hline
 $6$                 &  $38.04$    &             &             \\ \hline
 $7$                 &  $41.01$    &             &             \\ \hline
 $8$                 &  $53.56$    &             &             \\ \hline
 $9$                 &  $60.33$    &             &             \\ \hline
$10$                 &  $81.73$    &             &             \\ \hline
$11$                 &  $89.99$    &             &             \\ \hline
$12$                 & $156.64$    &             &             \\ \hline
$13$                 & $164.48$    &             &             \\ \hline
$14$                 & $188.93$    &             &             \\ \hline
$15$                 & $188.52$    &             &             \\ \hline
\end{tabular}
\begin{tabular}{|c|c|c|}\hline
\multicolumn{3}{|c|}{Lev+Uniform}       \\ \hline
new / ideal &  fb / ideal & f / ideal   \\ \hline
 $10.68$    &   $1686.54$ &  $74164.46$ \\ \hline
 $12.14$    &   $3392.04$ & $262620.28$ \\ \hline
 $14.59$    &  $73079.20$ &             \\ \hline
 $17.33$    & $136324.69$ &             \\ \hline
 $25.22$    &             &             \\ \hline
 $27.06$    &             &             \\ \hline
 $33.69$    &             &             \\ \hline
 $39.79$    &             &             \\ \hline
 $56.57$    &             &             \\ \hline
 $61.09$    &             &             \\ \hline
$114.70$    &             &             \\ \hline
$124.96$    &             &             \\ \hline
$137.76$    &             &             \\ \hline
$136.48$    &             &             \\ \hline
\end{tabular}
}
}
\caption{Comparison of search times for different symbol distributions. Randomly generated lexica.
f - Oflazer's method, fb - the forward-backward method, new - the new method, ideal - the ``ideal method''. All entries represent factors, comparing the given method with the ideal method.
Empty cells mean that we did not wait for termination.}\label{tableUniform}
\end{table}

The differences between the search times for the three methods for approximate search observed in Table~\ref{tableBasic} basically remain unchanged.
For bound $b=2$, search in the natural language lexicon of titles (Table~\ref{tableBasic}) is faster than search in the lexica with binomial distribution and is slower than search in the lexica with uniform distribution.  For larger distance bounds, the differences between the search times for the three types of lexica are more difficult to interpret.

\subsection{Influence of the length of the strings in the lexicon} 

In our last experiment we look at the influence of the length of the strings in the lexicon. 
We selected a smaller dictionary of natural language expressions consisting of approximately $450,000$ Bulgarian word forms with
average word length $10.01$. In Table~\ref{tableSmaller} the corresponding search times for the smaller dictionary
are found in columns $2-4$. Since for every query string $P$ we require $\vert P\vert \geq 3b$, for bounds $b > 4$
there are less than $10,000$ entries in the smaller dictionary from which we could generate queries. For this reason in the case of the smaller dictionary
we do not present results for $b > 4$. Even for the short strings of the Bulgarian lexicon, the new method is much faster than the forward-backward method and the third method.
The speed-up gained is less drastic than for the lexicons of titles and MEDLINE sentences, and here the ``ideal method'' remains more than $85$ times faster than the
new method.


\begin{table}
{\scriptsize
\noindent\makebox[\textwidth]{
\begin{tabular}{|c|c|c|c|}\hline
\multirow{2}{*}{$b$} & \multicolumn{3}{c|}{Levenshtein distance, lexicon of Bulgarian word forms} \\ \cline{2-4}
                     & new / ideal &  fb / ideal & f / ideal  \\ \hline
 $2$                 &  $85.86$    &  $207.10$   &  $1980.55$ \\ \hline
 $3$                 &  $105.28$   &  $510.04$   &  $9236.49$ \\ \hline
 $4$                 &  $420.32$   & $1905.09$   &            \\ \hline
\end{tabular}
}
}
\caption{Search times for Bulgarian lexicon with short strings showing the influence of the length of lexicon entries. Standard Levenshtein distance, search times for three methods.
f - Oflazer's method, fb - the forward-backward method, new - the new method, ideal - the ``ideal method''.}\label{tableSmaller}
\end{table}

{\em Size of index structures.\ } Table~\ref{tableSizes} represents for every method, except the ideal one,
the sizes in megabytes of the indexes compiled from the dictionary of titles, the dictionary of MEDLINE sentences and
the dictionary of Bulgarian word forms.

\begin{table}
{\scriptsize
\begin{center}
\begin{tabular}{|c|c|c|c|}\hline
method        & new       & fb       & f       \\ \hline
Titles        & $1104.63$ & $110.32$ & $54.18$ \\ \hline
MEDLINE       &  $921.41$ & $100.02$ & $49.87$ \\ \hline
Bg word forms & $61.02$   & $9.22$   & $4.17$  \\ \hline
\end{tabular}
\end{center}
}
\caption{Dictionary of titles and dictionary of MEDLINE sentences vs. smaller dictionary, sizes of indexes in megabytes.}\label{tableSizes}
\end{table}

\section{Historical remarks, possible applications and conclusion}\label{sec:conclusion}

We introduced a new method for fast approximate search in lexica that can be used for a large family of string distances.
The method uses a bidirectional index structure for the lexicon. This index structure can be
seen as a part of a longer development of related index structures starting with
work on suffix tries, suffix trees, and directed acyclic word graphs (DAWGs) \cite{Weiner73,McCreight76,Ukkonen95,ChenSeiferas84,BlumerBlumer85}.
These index structures address single texts and are
one-directional in the sense that search for substrings of the given string/text
follows the left-to-right reading order. In 
\cite{Stoye95,Stoye04,Maass00,Inenaga01} it has been shown how to
obtain {\em bidirectional} index structures for strings/texts, supporting search for
substrings using both left-to-right and right-to-left reading order. One-directional
index structures for {\em sets of strings} (as opposed to single strings) have been
described in \cite{BlumerBlumer87,Gusfield97,Breslauer98,MMW09,Inenaga05}. In each case the challenge is to find an index structure
with size linear in the size of the input
text or lexicon, with a linear-time construction algorithm. In \cite{BlumerBlumer87}
a {\em bidirectional index structure for sets of texts} is briefly sketched, asking
for natural applications. In this paper we have seen that such an index applied to
lexica can be used to realize a very fast method for approximate search.

With the new index, the ``wall effect'' mentioned in the Introduction can be avoided. Among related techniques, the  
BLASTA method \cite{AGMML90} is worth mentioning. In this approach, the occurrences of specific substrings in the lexicon are indexed in order to reduce the lexicon words to be considered. It assumes that each answer of the query has to contain at least one of the keyed substrings which allows it to start with an exact match of such a promising substring. In such a way BLASTA prunes the initial exhaustive search and proves to be efficient. However since
there is no guarantee that all answers of the query meet this condition, it may fail to retrieve the complete list of words satisfying the query.

Our evaluation results show that the new method is much faster than
previous methods, and for lexica with long strings the speed-up is drastic. 
Here the new method for distance bound $b=2$ comes close to the theoretical limit when using the standard Levenshtein distance.

We add a brief comment on possible applications. As a matter of fact, the method may be used to speed-up traditional spelling correction techniques. For high quality spelling correction, speed is not the only issue. Current approaches typically use probabilistic techniques at two places. First, good similarity measures for selecting candidates are based on special edit operations with weights depending on the particular symbols/strings used. How to find appropriate edit operations and weights is a question beyond the scope of the current paper. However, the framework of a generalized distance
we use to model similarities should be general enough to cover most interesting cases. Second, when looking for an optimal correction suggestion for a misspelled token, language models (e.g., weighted word trigrams) help to find a correction suggestion that fits the local context. Still, {\em similarity search} in the background lexicon only looks at {\em single} tokens and for efficiency reasons, ``context sensitive'' correction suggestions for distinct tokens are often computed in isolation. An interesting question is if better results are obtained when using larger contexts already for the background lexica and similarity search. This strategy would guarantee that the correction suggestions obtained for a sequence of tokens always fit together. The method introduced above offers new possibilities for testing such a strategy since we can use large strings and distance bounds. As a matter of fact, issues of smoothing have to be taken into account when trying to synchronize contextual similarity search and language models.  

Possible application areas of the new method are not restricted to traditional fields of approximate search such as spell checking, text and OCR correction. Since the method is fast enough to deal with  collections of long strings and large distance bounds, it seems promising to test its use, e.g., for detecting plagiarism, for finding similar sentences in translation memories and related language databases, and for approximate search in collections of address or bibliographic data. 
We currently also look at a variant of the method for fast approximate search of patterns in an indexed collection of texts. In order to find all approximate matches for a string in a (collection of) texts, the index has to be enriched by adding information on the positions of all occurrences of each infix. The challenge is to keep the size of the index linear in the size of the text(s).

A remaining open question is the time complexity of the presented algorithm. A desirable approach would be to estimate the average complexity in a way similar to Myers' \cite{M94}.

There are two obvious ways how search times presented above could be immediately improved.
First, for small distance bound we could use universal Levenshtein automata \cite{MS04} as filters. This leads to a performance gain as compared to
matrix based filters \cite{MitankinMihovSchulzTCS2011}. Second, an additional speed-up could be obtained by running subsearches
of distinct branches of the search tree used in parallel. The optimal selection of search trees is an interesting point for further investigations.

A method for approximate search for Hamming distance utilizing a bidirectional index structure is presented in~\cite{LLT09}. 
The search method is presented only for error bounds $\leq 2$ and no generalization for higher error bounds and/or other distances is given. 
The bidirectional index structure in~\cite{LLT09} is based on compressed suffix array. 
In this way one can significantly reduce the space required for the index structure. 
However, the cost of a single transition increases to $O(\log |\Sigma|/ \log \log ||\mathcal{D}|| + 1)$, 
as suggested in \cite{Hoang12} page 74, compared to $O(1)$ in the SCDAWG. 
Clearly our search method can be applied with the bidirectional index structures presented in~\cite{LLT09} and in~\cite{Schnattinger2012}.

\bibliographystyle{alpha}
\bibliography{BibExample}

\section*{Appendix}

We show how the search strategy described in Section~\ref{sec:searching} can be adapted to the
case of an arbitrary generalized distance $d=(\Op,w)$. In what follows,
$\omega_{\max}$ denotes the maximal width of an operation $\op\in \Op$.
To simplify the following description, we introduce the notion of a (left, right) reduct of a word. Intuitively, reducts of a word $U$ are obtained by deleting a ``short''
(possibly empty) prefix and/or suffix of length $< \omega_{max}$ from $U$.
\begin{definition}
Let $U\in \Sigma^\ast$ be represented in the form $U=U_1\circ U_2$.
If $\vert U_1\vert < \omega_{max}$, then $U_2$ is called a {\em left reduct} of $U$.
If $\vert U_2\vert < \omega_{max}$, then $U_1$ is called a {\em right reduct} of $U$.
If $U=U_1\circ V \circ U_2$ and both $\vert U_1\vert < \omega_{max}$ and $\vert U_2\vert < \omega_{max}$, then $V$ is called a {\em reduct} of $U$.
\end{definition}
We denote that for $\omega_{max}=1$ always $U$ is the only reduct of $U$.
The formal background for the adapted search procedure is provided by the following generalization of
Proposition~\ref{PropositionSplitAlternative}.

\begin{proposition}\label{PropositionSplitGeneralised}
Let $P'=P'_1\circ P'_2$ and $\alpha$ be an alignment with $l(\alpha)=P'$ and $w(\alpha) \leq b'$, then:
\begin{enumerate}
\item
$\alpha$ can be represented in the form $\alpha=\alpha_1\circ \beta\circ \alpha_2$ such that
$\beta\in \Op \cup \{\varepsilon\}$, $l(\alpha_1)$ is a right reduct of $P_1'$, and
$l(\alpha_2)$ is a left reduct of $P_1'$.
\item
for each such decomposition and integers $b'_1$ and $b'_2$ with $b'_1+b'_2=b'-1$
it holds that $w(\alpha_1)\le b'_1$ or $w(\alpha_2)\le b'_2$.
\end{enumerate}
\end{proposition}

{\em Proof.\ }
We first prove Part~1.
Let $\alpha_1$ denote the maximal prefix of $\alpha$ with the property that
$l(\alpha_1)$ is a prefix of $P'_1$. If $l(\alpha_1)=P'_1$ we define
$\beta := \varepsilon$. Otherwise there exists an operation $\op\in \Op$ such that
$l(\alpha_1)$ is a proper prefix of $P'_1$,  the latter being a proper prefix of $l(\alpha_1\circ \op)$.
In this case we define $\beta := \op$. In both cases $\alpha_2$ is now determined by the equation $\alpha=\alpha_1\circ \beta\circ \alpha_2$.
It is trivial to check that this representation has the properties stated above.
%
%
The second statement follows easily.
\qed

Recall that in the special situation considered in Section~\ref{sec:searching} we decomposed the pattern $P$ into subparts $P=P_1\circ P_2\cdots \circ P_{b+1}$, and for substrings
of the form $P_k \circ \cdots \circ P_l$ ($k \leq l$, possible combinations of $k, l$ determined by the structure of the search tree) we
computed approximate matches with substrings of lexicon words using distinct bounds. In the general situation considered here
we split $P$ as above. We then try to find approximate matches between {\em reducts} of the substrings $P_k \circ \cdots \circ P_l$ with substrings of lexicon words.
For a formal description, let us introduce another notational convention.
By $r(i,U,j)$ we denote the reduct obtained from $U$ by deleting the unique prefix and suffix of length $i$ and $j$, respectively. Hence $r(0,U,0)= U$.

\noindent{\bf Building the generalized search tree for a pattern.\ } For a given input pattern $P$ and a bound $b$, let ${\mathcal T}_P$ denote the search tree defined in Section~\ref{sec:searching}.
With each query $(P',b')$ decorating a node $\eta$ we associate as a subcase analysis the set of all
{\em derived} queries of the form $(r(i,P',j),b')$ where $i,j< \omega_{max}$. The problem considered at node
$\eta$ is to solve all derived queries of the above form. Note that $(r(0,P,0),b)$ is equivalent to $(P,b)$.

\noindent{\bf Computation of solution sets for derived queries.\ }
For each derived query $(r(i,P',j),b')$ of the generalized tree we compute a set $S_D(r(i,P',j),b')$ in a bottom-up fashion.
We shall
prove below that $S_D(r(i,P',j),b')$ is the solution set $\Sol_D(r(i,P',j),b')$ in each case.

{\em Initialization steps.\ } For a derived query $(r(i,P',j),0)$ at a leaf
we decide if $r(i,P',j)$ is a substring of a lexicon word. In the positive case we let $S_D(r(i,P',j),0) := \{r(i,P',j)\}$, otherwise we define $S_D(r(i,P',j),0) := \emptyset$.

{\em Extension steps.\ }
Let $((r(i,P',j),b')$ denote a derived query at a non-leaf node $\eta$ of ${\mathcal T}_P$, let
let $(P_1',b_1')$ and $(P_2',b_2')$ denote the main queries of the two children $\eta_1, \eta_2$ of $\eta$, which are given in the natural left-to-right ordering.
Given all sets $S_D((r(i,P'_1,j_1),b_1')$ and $S_D((r(i_2,P'_2,j),b_2')$ for the derived queries at $\eta_1, \eta_2$ we define $S_D(r(i,P',j),b')$ as the union of the two sets
$S_1$ and $S_2$ defined as
\begin{eqnarray*}
S_1&=&\bigcup_{j_1=0}^{\omega_{max}-1}\{U \circ V\in \Inf({\mathcal{D}}) \mid U\in S_D(r(i,P'_1,j_1),b_1'), d_1^\ast\le b'\}\\
S_2&=&\bigcup_{i_2=0}^{\omega_{max}-1}\{V \circ U\in \Inf({\mathcal{D}}) \mid U\in S_D(r(i_2,P'_2,j),b_2'), d_2^\ast\le b'\}\}.
\end{eqnarray*}
Here $d_1^\ast = d(r(i_1,P'_1,j_1),U)+d(Q_1,V)$ where $Q_1$ is obtained from $P'=P_1'P_2'$ by deleting the prefix
of length $\vert P_1'\vert  - j_1$ and the suffix of length $j$.
Similarly $d_2^\ast= d(r(i_2,P'_2,j),U)+d(Q_2,V)$ where $Q_2$ is obtained from $P'$ by deleting the prefix of length $i$ and the suffix of length $|P_2'| - i_2$.

\begin{proposition}\label{GeneralizedPropositionCorrectness}
The computation of solution sets is correct: for each derived query $(r(i,P',j),b')$ we have $S_D(r(i,P',j),b')=\Sol_D(r(i,P',j),b')$.
\end{proposition}

The proof is a simple modification of the earlier correctness proof.
We just use Proposition~\ref{PropositionSplitGeneralised} instead of Proposition~\ref{PropositionSplitAlternative}.

\end{document}